%% file: main.tex
\documentclass[10pt,twocolumn,letterpaper]{article}

\usepackage{cvpr}              

\definecolor{cvprblue}{rgb}{0.21,0.49,0.74}
\usepackage[pagebackref,breaklinks,colorlinks,allcolors=cvprblue]{hyperref}
\usepackage[accsupp]{axessibility}

\usepackage{graphicx}
\usepackage{algorithm}
\usepackage{algpseudocode}
\usepackage{booktabs}
\usepackage{multirow}
\usepackage{color}
\usepackage[table]{xcolor}
\usepackage{listings}
\usepackage{wrapfig}
\usepackage{enumitem}
\usepackage{adjustbox}
\usepackage[most]{tcolorbox}
\usepackage{hyperref}
\usepackage{url}
\input{preamble}

\newtcolorbox{codebox}[1][]{%
  breakable,
  enhanced,
  colback=gray!3,
  colframe=gray!50!black,
  arc=2mm,
  boxrule=0.5pt,
  left=2mm,
  right=2mm,
  fontupper=\ttfamily\small,
  listing only,
  listing options={
    basicstyle=\ttfamily\small,
    breaklines=true,
    columns=fullflexible,
    keepspaces=true,
    mathescape=true,
  },
  #1
}

\newtcolorbox{responsebox}[1][]{%
  breakable,
  enhanced,
  colback=cyan!4,
  colframe=gray!50!black,
  arc=2mm,
  boxrule=0.5pt,
  left=2mm,
  right=2mm,
  fontupper=\ttfamily\small,
  #1
}

\def\paperID{} 
\def\confName{CVPR}
\def\confYear{2026}

\title{CoLLM-NAS: Collaborative Large Language Models for Efficient Knowledge-Guided Neural Architecture Search}

\author{
Zhe Li \quad 
Zhiwei Lin \quad 
Yongtao Wang \thanks{Corresponding author.} \\
Wangxuan Institute of Computer Technology, Peking University \\
{\tt\small zheli@stu.pku.edu.cn \quad \{zwlin, wyt\}@pku.edu.cn}
}

\begin{document}
\maketitle
\input{sections/00_abstract}    
\input{sections/01_introduction}
\input{sections/02_related_work}
\input{sections/03_method}
\input{sections/04_experiment}
\input{sections/05_conclusion}
\input{sections/06_acknowledgement}
{
    \small
    \bibliographystyle{ieeenat_fullname}
    \bibliography{main}
}

\clearpage
\onecolumn  
\setcounter{page}{1}

\renewcommand{\maketitlesupplementary}{%
   \newpage
   \begin{center}
   \Large
   \textbf{\thetitle}\\
   \vspace{0.5em}Supplementary Material \\
   \vspace{1.0em}
   \end{center}
}

\maketitlesupplementary
\appendix
\input{appendix/00_macro_search_spaces}
\input{appendix/01_temperature}
\input{appendix/02_PoC_experiment}
\input{appendix/03_ofa_acc_predictor}
\input{appendix/04_prompts}
\input{appendix/05_architectures}
\end{document}

%% file: preamble.tex
\def\ours{CoLLM-NAS}
\definecolor{Gray}{gray}{0.9}
\definecolor{Blue}{rgb}{0.92, 0.96, 1.0}
\definecolor{LightRed}{rgb}{0.8, 0.3, 0.3}
\definecolor{LightOrange}{rgb}{0.99, 0.92, 0.88}

%% file: sections/00_abstract.tex
\begin{abstract}
The integration of Large Language Models (LLMs) with Neural Architecture Search (NAS) has introduced new possibilities for automating the design of neural architectures. 
However, most existing methods face critical limitations, including architectural invalidity, computational inefficiency, and inferior performance compared to traditional NAS.
In this work, we present \textbf{Co}llaborative \textbf{LLM}-based \textbf{NAS} (\ours), a two-stage NAS framework with knowledge-guided search driven by two complementary LLMs. 
Specifically, we propose a stateful \textit{Navigator LLM} to guide search direction, a stateless \textit{Generator LLM} to synthesize high-quality candidates, and a \textit{Coordinator} module to orchestrate inter-LLM communication and manage evaluation processes.
\ours~efficiently guides the search process by combining LLMs' inherent knowledge of structured neural architectures with progressive knowledge from iterative feedback and historical trajectory. 
Experimental results on ImageNet and NAS-Bench-201 show that \ours~surpasses existing NAS methods and conventional search algorithms, achieving new state-of-the-art results while significantly reducing search costs by 4--10$\times$.
Furthermore, \ours~consistently enhances the performance and efficiency of various two-stage NAS methods (\textit{e.g.,} OFA, SPOS, and AutoFormer) across diverse search spaces (\textit{e.g.,} MobileNet, ShuffleNet, and AutoFormer), demonstrating its excellent generalization.

\end{abstract}

%% file: sections/01_introduction.tex
\section{Introduction}
\label{sec:introduction}

\begin{figure}[!t]
    \centering
    \includegraphics[width=1\linewidth]{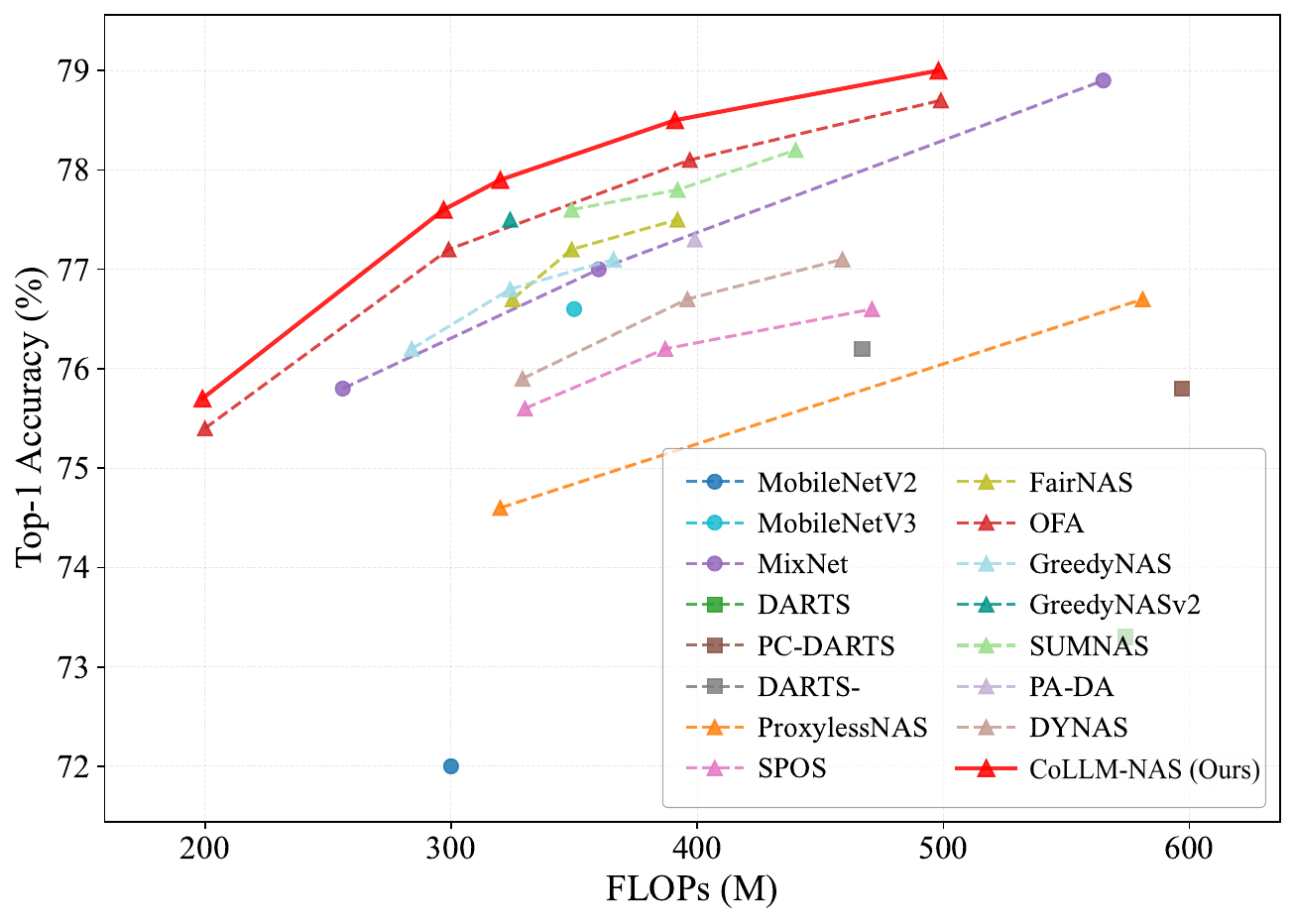}
    \caption{FLOPs-accuracy trade-off of architectures discovered by different NAS methods on ImageNet.}
    \label{fig:tradeoff}
\end{figure}

Designing neural architectures remains a critical task in the era of deep learning.
Neural Architecture Search (NAS), a cornerstone of AutoML, is dedicated to automating the discovery of optimal neural architectures. 
Early NAS methods typically employ reinforcement learning \cite{zoph2017neural, zoph2018learning, tan2019mnasnet} or evolutionary algorithms \cite{real2017large, real2019regularized} to search for high-performing architectures. 
While effective, these methods require training numerous independent architectures from scratch, resulting in prohibitive computational costs. 
%
To address this issue, one-shot NAS \cite{liu2018darts, guo2020single} is developed, utilizing a weight-sharing supernet to amortize training expenses. 
Two-stage NAS \cite{guo2020single, cai2020onceforall, AutoFormer} further decouples this process into two phases, \textit{i.e.,} supernet training and architecture search. 
Although this approach avoids costly retraining by inheriting supernet weights, conventional search algorithms (\textit{e.g.,} evolutionary algorithms) still require sampling and evaluating thousands of candidates in vast search spaces during the second phase to find optimal architectures, incurring notable test costs and risking convergence to local optima.

Recently, large language models (LLMs) have emerged as novel participants in the NAS landscape, leveraging their powerful representation capabilities, code generation proficiency, and reasoning abilities to enhance the search process from diverse perspectives \cite{wu2024evolutionary}. 
However, most existing LLM-based NAS methods \cite{chen2023evoprompting, nasir2024llmatic} directly modify architectures at the code level within unconstrained programming-language token spaces, often resulting in architectural invalidity, limited robustness, and independent training of each candidate.
Consequently, these methods fail to surpass established NAS baselines on standard benchmarks \cite{deng2009imagenet,dong2020nasbench201} while generating substantial computational overhead \cite{nasir2024llmatic}.
To address these limitations, we propose integrating the advantages of LLMs with mature two-stage NAS methods to provide more effective and efficient neural architecture design.
%
Specifically, we aim to enhance the search phase of two-stage NAS by replacing conventional search algorithms with knowledge-guided LLM reasoning, enabling intelligent navigation of the search space to accelerate convergence to high-performing regions.

To this end, this paper introduces a novel \textbf{Co}llaborative \textbf{LLM}-based \textbf{NAS} framework (CoLLM-NAS) that leverages synergistic interactions between two complementary LLMs to efficiently discover high-performing architectures for two-stage NAS. 
Concretely, we design three key components: 
(1) A stateful \textit{Navigator LLM} that provides search strategies through dynamic refinement, leveraging iterative evaluation feedback and historical trajectory analysis. 
(2) A stateless \textit{Generator LLM} that synthesizes high-quality architectures according to these strategies.
(3) A \textit{Coordinator} module that orchestrates inter-LLM communication, validates architectural legality, evaluates candidate performance, and maintains an architecture archive.
At its core, \ours~employs a collaborative framework where the \textit{Navigator} and \textit{Generator LLMs} work in concert, guiding the search with dual knowledge sources, {\textit{i.e.,}} LLMs' inherent architectural prior and progressive insights learned from iterative feedback and historical trajectory. The stateful-stateless design further enhances the exploration-exploitation balance.
Extensive experimental results demonstrate that our method can be applied to various two-stage NAS methods across diverse search spaces, consistently outperforming baselines under different resource constraints while significantly reducing search costs by 4--10$\times$. As shown in \cref{fig:tradeoff}, architectures discovered by \ours~achieve superior FLOPs-accuracy trade-off compared to those from state-of-the-art NAS methods on ImageNet.

The main contributions of this work are as follows:
\begin{itemize}
\item We present a novel collaborative LLM-based NAS framework, \ours, to enhance two-stage NAS with knowledge-guided search. To our knowledge, this is the first work integrating LLMs with two-stage NAS.
\item We propose three key components: a stateful \textit{Navigator LLM} to provide adaptive search strategies, a stateless \textit{Generator LLM} to synthesize high-quality architectures, and a \textit{Coordinator} module to orchestrate LLMs' interactions and manage evaluation processes.
\item Experiments demonstrate that \ours~surpasses existing NAS methods across diverse search spaces both in performance and efficiency, achieving new SOTA results while reducing search costs significantly.
\end{itemize}

%% file: sections/02_related_work.tex
\section{Related Work}
\label{sec:related-work}

\subsection{Two-stage NAS}
\label{sec:related-work:two-stage-nas}
Two-stage NAS, a prominent branch of one-shot NAS, addresses the computational inefficiency of traditional NAS by employing weight-sharing mechanisms. This approach decomposes the NAS problem into two sequential phases:
\begin{equation}
    \label{eq:two-stage-nas}
    \begin{aligned}
    w_{\mathcal{A}}^* &= \underset{w_{\mathcal{A}}}{\arg\min}\, \mathbb{E}_{\alpha \sim \Omega(\mathcal{A})}\left[\mathcal{L}(w_{\mathcal{A}}(\alpha), \mathcal{D}^{\text{train}})\right], \\
    \alpha^* &= \underset{\alpha \in \mathcal{A}}{\arg\max}\, \mathcal{P}(w_{\mathcal{A}}^*(\alpha), \mathcal{D}^{\text{val}}) \quad \text{s.t.}  \text{Cost}(\alpha) \leq \Lambda,
    \end{aligned}
\end{equation}
where $\mathcal{A}$ denotes the search space, $w_{\mathcal{A}}$ represents shared supernet weights, $w_{\mathcal{A}}(\alpha)$ indicates subnet weights inherited from the supernet, $\Omega(\mathcal{A})$ represents the sampling strategy, $\mathcal{P}$ and $\Lambda$ are the performance evaluation function and the resource constraint, respectively.
By decoupling architecture parameter optimization from network weight training, it eliminates conflicts in simultaneous updates and reduces mutual interference. 

In the first stage, a weight-sharing supernet is trained via diverse sampling strategies. SPOS \cite{guo2020single} pioneers uniform single-path sampling.
OFA \cite{cai2020onceforall} enables multi-scale subnet extraction via progressive shrinking. In \cite{lu2023pa}, Pa\&da jointly samples paths and data for consistent training. DYNAS \cite{jeon2025subnet} employs subnet-aware dynamic supernet training strategy. While AutoFormer \cite{AutoFormer} adapts this weight-sharing mechanism for vision transformers.
In the second stage, optimal architectures are searched under resource constraints. During evaluation, candidates can directly inherit weights from the supernet, enabling efficient performance estimation without retraining. Since exhaustive evaluation remains infeasible, most methods employ random search \cite{li2020random}, reinforcement learning \cite{zoph2017neural}, or evolutionary algorithms \cite{guo2020single, cai2020onceforall} to efficiently explore the search space.

This paper introduces an LLM-based knowledge-guided search paradigm to replace conventional search algorithms in the second stage, aiming to significantly reduce evaluations and achieve better performance with lower costs.

\subsection{LLM for NAS}
\label{sec:related-work:llm-for-nas}
The rise of LLMs offers a fundamentally new perspective for NAS by reformulating the search process through advanced reasoning capabilities and semantic understanding.
GENIUS \cite{zheng2023can} pioneers GPT-4 as a black-box optimizer to generate and refine architectures via natural language. EvoPrompting \cite{chen2023evoprompting} repositions LLMs as mutation and crossover operators, employing evolutionary prompt engineering and soft prompt-tuning to explore a vast code-based search space. Similarly, LLMatic \cite{nasir2024llmatic} leverages LLMs as operators but uniquely integrates Quality-Diversity Optimization to create diverse networks. A reflective zero-cost strategy is developed by RZ-NAS \cite{ji2025rznas}, combining LLM reflection modules with training-free metrics. LM-Searcher \cite{hu2025lm} reformulates NAS as a ranking task using NCode, enabling cross-domain architecture search. While NADER \cite{yang2025nader} employs a multi-agent approach for neural architecture design, its open design space and full training requirements limit its scalability to ImageNet-scale datasets.

In contrast, this paper proposes a collaborative LLM-based NAS framework with knowledge-guided search. Moreover, our approach leverages a pre-trained supernet for efficient evaluation and inherits its pre-optimized search space to avoid the invalidity issue of generated architectures, enabling scalable search on large-scale datasets.

%% file: sections/03_method.tex
\section{Method}
\label{sec:method}
In this section, we first validate LLMs' capability to comprehend structured neural architectures within hand-crafted search spaces. Building on this, we introduce \ours, our collaborative LLM-based NAS framework.

\subsection{Architecture Comprehension in LLMs}
\label{sec:method:arch-comprehension-in-llms}

\begin{figure}[!t]
    \centering
    \begin{tabular}{@{}c@{}c@{}}
        \includegraphics[width=0.5\linewidth]{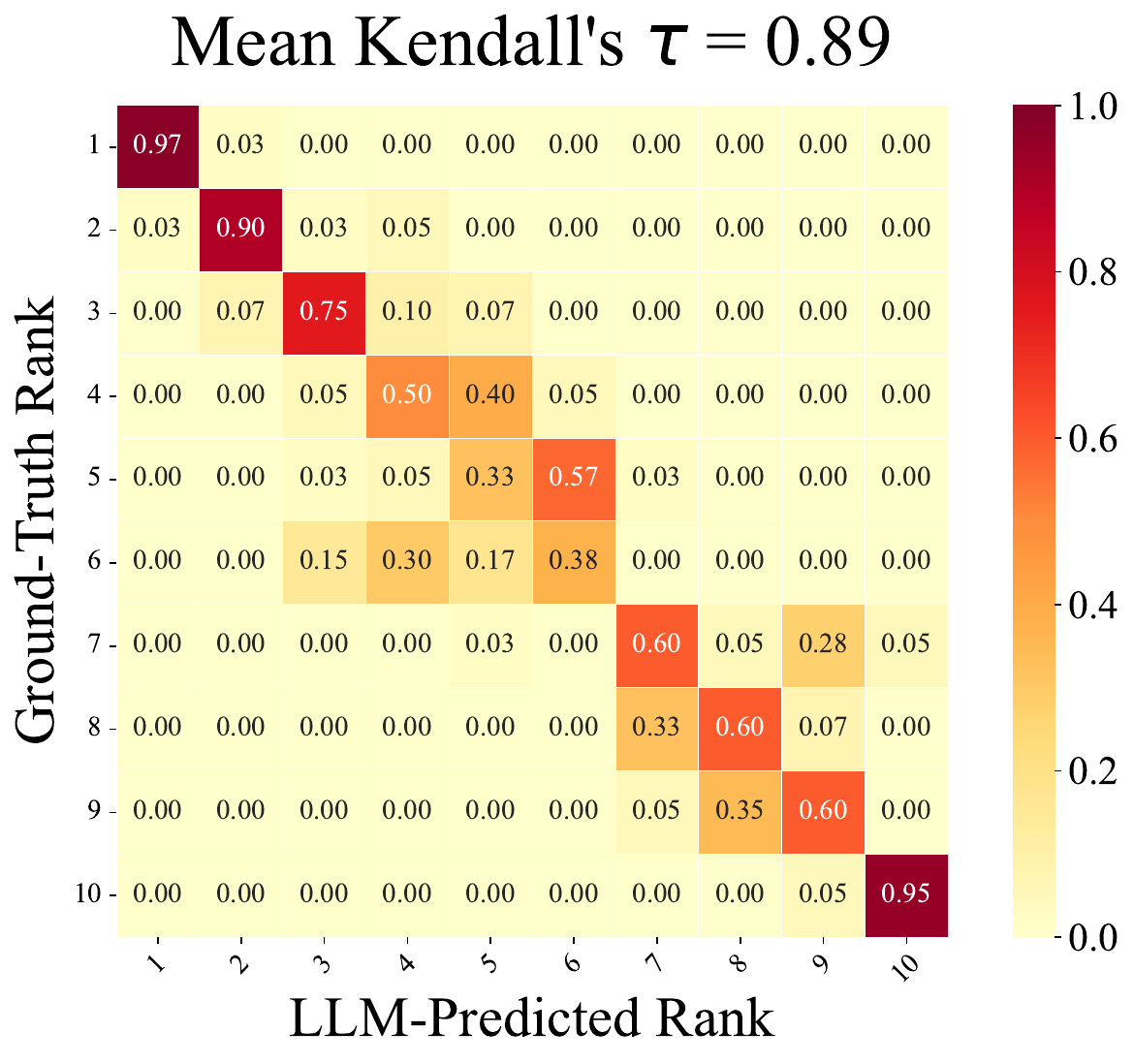} &
        \includegraphics[width=0.5\linewidth]{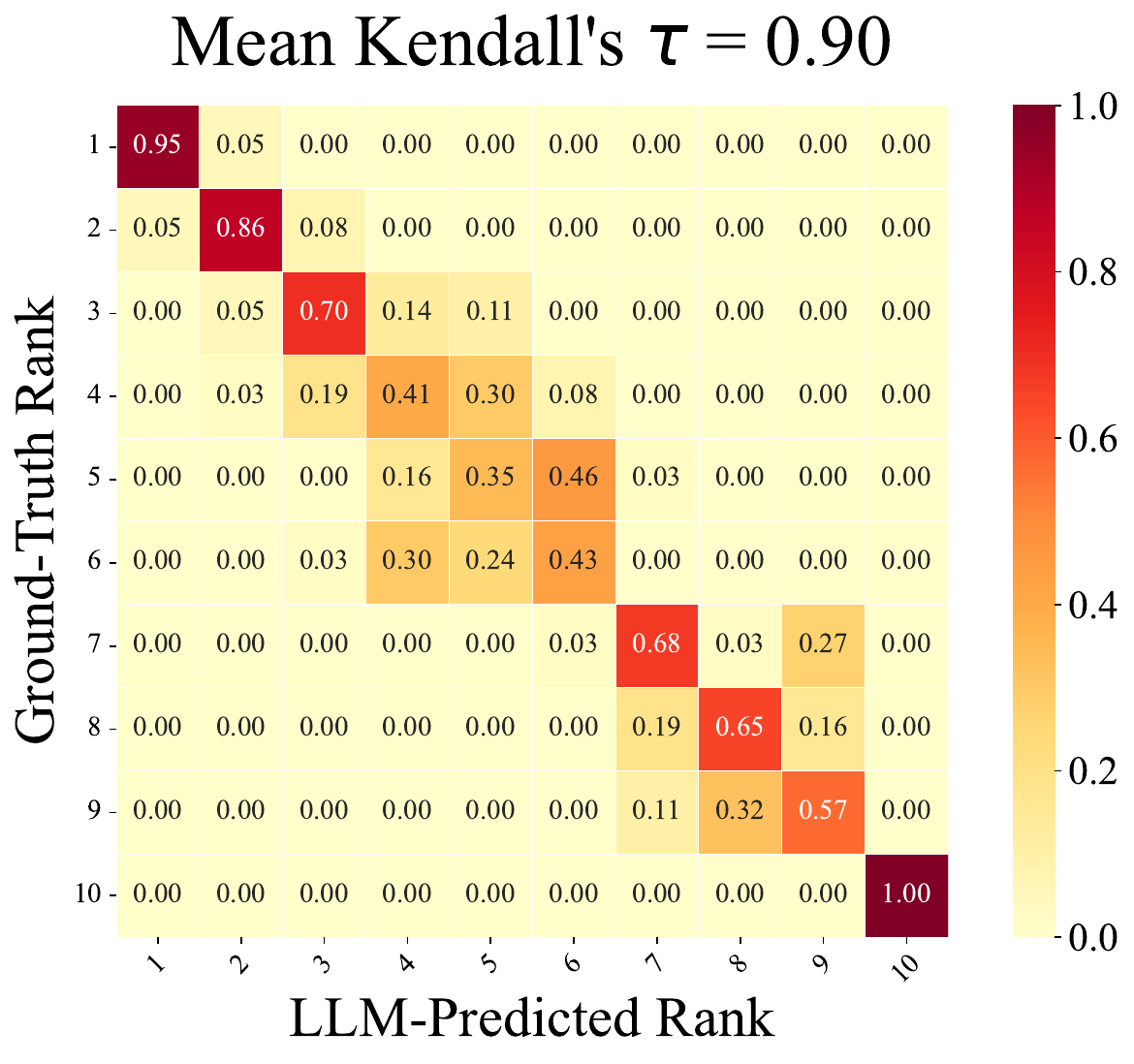} \\
        (a) CIFAR-10 & (b) CIFAR-100
    \end{tabular}
    \caption{Consistency heatmap between LLM-predicted and ground-truth rankings on CIFAR-10 and CIFAR-100 within NAS-Bench-201 search space.}
    \label{fig:ranking_heatmap}
\end{figure}

Recent advances in LLMs demonstrate their capability to understand complex technical domains, acquired through pre-training on extensive technical literature. We hypothesize that \textit{modern LLMs have internalized knowledge of neural architecture design principles that could be valuable for NAS.} To investigate this capability, we design an experiment on NAS-Bench-201.

\paragraph{Proof-of-Concept Experiment.}
We partition architectures in NAS-Bench-201 into 10 equal-sized subsets by their precomputed performance, and sample one architecture from each subset.
The Qwen3-30B-A3B LLM \cite{qwen3technicalreport} is prompted to rank these architectures based on its comprehension of neural network design principles, while being blinded to ground-truth accuracy. 
Architectures are represented in their standard graph-based encoding format.
To ensure statistical robustness, we perform 40 independent trials on CIFAR-10 and CIFAR-100.
%
As illustrated in \cref{fig:ranking_heatmap}, results reveal strong alignment between LLM predictions and empirical performance, achieving mean Kendall's $\tau$ values of 0.89 on CIFAR-10 and 0.90 on CIFAR-100. Moreover, the best architecture is correctly identified in the vast majority of trials.
This demonstrates that \textit{LLMs exhibit non-trivial comprehension of neural architecture performance patterns}, even when operating on structured representations within hand-crafted search spaces.
Notably, modern LLMs may have encountered some architectures of NAS-Bench-201 in their training corpus.
To avoid prior knowledge contamination, our prompts prevent transmission of any explicit information about the benchmark. 
Furthermore, as evident from the reasoning contents in Appendix C, the LLM's ranking decisions stem from its understanding of architectural principles (\textit{e.g.,} operator effectiveness and information flow) rather than memorized architecture-performance mappings.


\subsection{\ours~Framework}
\label{sec:method:collm-nas}

\begin{algorithm}[!t]
    \small
    \caption{\ours: Exploring High-Performing Architectures via Collaborative LLMs}
    \label{alg:collm-nas-alg}
    \begin{algorithmic}[1]
    \Require Target accuracy $P_{target}$, Resource constraint $\Lambda$, Iteration limit $T$
    \Ensure Best architecture $\alpha^*$
    
    \State \textbf{Initialization:} $\alpha^* \gets \emptyset$, $p^* \gets 0$, $\mathcal{V} \gets \emptyset$, $\mathcal{H} \gets \emptyset$ \\
    \Comment{Best arch, best accuracy, visited set, and history}
    \State $\mathcal{S}_0 \gets \textsc{NavigatorLLM}(P_{target}, \Lambda)$
    \Comment{Initialize strategy}
    
    \For{$t = 1$ \textbf{to} $T$}
        \State $\mathcal{C}_t \gets \textsc{GeneratorLLM}(\mathcal{S}_{t-1})$
        \Comment{Generate candidates}
        \State $\mathcal{R}_t \gets \emptyset$ 
        \Comment{Collection of evaluations}
        
        \For{each $\alpha_i \in \mathcal{C}_t \setminus \mathcal{V}$ \textbf{such that} $\textsc{isLegal}(\alpha_i)$}
            \State Compute cost $c_i$ and evaluate performance $p_i$ for $\alpha_i$
            \State $\mathcal{R}_t \gets \mathcal{R}_t \cup \{(\alpha_i, p_i, c_i)\}$; $\mathcal{V} \gets \mathcal{V} \cup \{\alpha_i\}$
            \If{$c_i \leq \Lambda$ \textbf{and} $p_i > p^*$} $p^* \gets p_i$, $\alpha^* \gets \alpha_i$ \EndIf
        \EndFor
        
        \If{$p^* \geq P_{target}$} \textbf{break} \EndIf
        \State $\mathcal{H} \gets \mathcal{H} \cup \{(\mathcal{S}_{t-1}, \mathcal{R}_t)\}$ \Comment{Update history}
        \State $\mathcal{S}_t \gets \textsc{NavigatorLLM}(\mathcal{H})$ \Comment{Refine strategy}
    \EndFor
    \State \Return $\alpha^*$
    \end{algorithmic}
\end{algorithm}

\begin{figure*}[!t]
    \centering
    \includegraphics[width=0.97\linewidth]{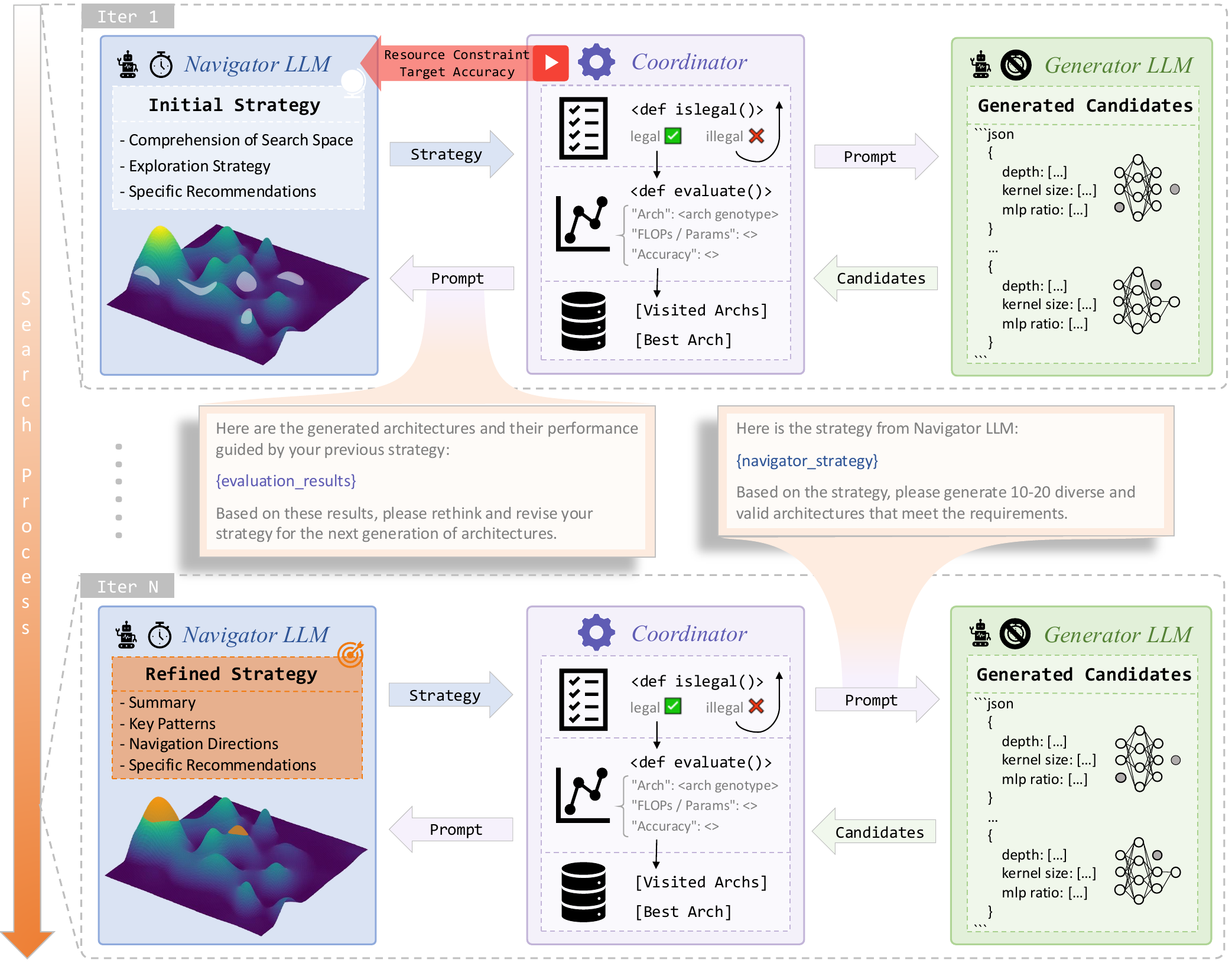}
    \caption{Pipeline of \ours. The search starts with the \textit{Navigator LLM} generating an initial exploration strategy based on target accuracy and resource constraint (\textit{e.g.,} FLOPs, Params). The \textit{Coordinator} then transmits this strategy to \textit{Generator LLM}, which synthesizes high-quality candidates accordingly. After \textit{Coordinator} validation and evaluation, results are fed back to \textit{Navigator LLM} for strategy refinement. This loop iterates until achieving the target accuracy or reaching the iteration limit. Orange regions indicate high-performing areas of search spaces.}
    \label{fig:pipeline}
\end{figure*}

As shown in \cref{fig:pipeline}, \ours~works through a collaboration process, consisting of three key components: a \textit{Navigator LLM}, a \textit{Generator LLM}, and a \textit{Coordinator} module.

\paragraph{Navigator LLM.}
The stateful \textit{Navigator LLM} functions as a strategic guide with persistent memory. 
Through iterative analysis of performance patterns emerging from evaluated architectures, it dynamically formulates and refines search strategies. These strategies progressively concentrate on high-potential regions of the search space. 
In the initial phase, the \textit{Navigator LLM} is prompted to establish an exploration strategy that promotes architectural diversity, improving initial population quality through its implicit comprehension of architectures. 
As iterations progress, it continuously refines this strategy based on accumulated feedback, transitioning from broad exploration to targeted exploitation of identified high-performing regions.

\vspace{-2mm}

\paragraph{Generator LLM.}
The stateless \textit{Generator LLM} serves as a specialized architecture synthesizer, focusing exclusively on the current strategy without retaining any memory. Following \textit{Navigator LLM}'s guidance, it translates the strategy into concrete candidate architectures during each iteration. These candidates simultaneously conform to the search space constraints while embodying the architectural patterns emphasized by the current strategy.

\paragraph{Coordinator.}
The \textit{Coordinator} manages the overall search process. It orchestrates inter-LLM communication, verifies architectural legality, evaluates candidate performance, and maintains an archive of visited architectures to eliminate redundant evaluations. 
Critically, the evaluation employs weight-sharing mechanisms and inherits weights from a pre-trained supernet, enabling rapid performance assessment while preserving learned parameter relationships.

Through system prompt design, we assign distinct roles and responsibilities to each LLM, inform them of the collaboration process, and convey knowledge about different search spaces and architectural representations. 
Similarly, to prevent knowledge contamination, we avoid transmitting explicit search space information in our prompts. Detailed examples of prompts and responses are shown in Appendix E. Our search workflow is described in \cref{alg:collm-nas-alg}. Through this collaborative approach, \ours~intelligently navigates search spaces and efficiently identifies high-performing architectures.

%% file: sections/04_experiment.tex
\section{Experiment}
\label{sec:experiment}

\subsection{Experimental Setup}
\label{sec:experiment:settings}
We evaluate the proposed \ours~in three macro search spaces: MobileNet \cite{cai2018proxylessnas}, ShuffleNet \cite{ma2018shufflenet}, and AutoFormer \cite{AutoFormer}, and one micro cell-based search space: NAS-Bench-201 \cite{dong2020nasbench201}.

\begin{table}[!ht]
    \caption{Key experimental settings. "\textit{same}" indicates identical settings to the corresponding baseline.}
    \label{tab:settings}
    \small
    \centering
    \begin{tabular}{l|cccc}
        \toprule
        \multicolumn{1}{l|}{Method} & 
        \multicolumn{1}{c}{Evaluation} & 
        \begin{tabular}[c]{@{}c@{}}Resource\\Constraint\end{tabular} & 
        \multicolumn{1}{c}{Retrain} \\
        \midrule
        OFA & predict$\rightarrow$validate & FLOPs & $\times$ \\
        SPOS & validate & FLOPs & $\checkmark$ \\
        AutoFormer & validate & Params. & $\times$ \\
        Ours & validate & \textit{same} & \textit{same} \\
        \bottomrule
    \end{tabular}
    \vspace{-4mm}
\end{table}

\paragraph{Macro Search Spaces.}

\begin{table}[!t]
    \caption{Performance comparison on ImageNet within macro search spaces. "GPU Days" only covers the search phase (excluding supernet training), and "Arch. Budget" refers to the budget on the number of explored architectures.}
    \label{tab:macro-search-space}
    \small
    \setlength{\tabcolsep}{1mm}
    \centering
    \begin{tabular}{l|c|c|c|c|c}
        \toprule
        \multicolumn{1}{l|}{Method} & 
        \begin{tabular}[c]{@{}c@{}}Top-1\\(\%)\end{tabular} & 
        \begin{tabular}[c]{@{}c@{}}FLOPs\\(M)\end{tabular} & 
        \begin{tabular}[c]{@{}c@{}}Params.\\(M)\end{tabular} & 
        \begin{tabular}[c]{@{}c@{}}GPU\\Days\end{tabular} & 
        \begin{tabular}[c]{@{}c@{}}Arch.\\Budget\end{tabular} \\
        \midrule
        \multicolumn{6}{c}{MobileNet Search Space} \\
        \midrule
        OFA-T & 75.4 & 200 & 3.6 & \multirow{4}{*}{0.42} & \multirow{4}{*}{1000} \\
        OFA-S & 77.2 & 299 & 4.2 &  &  \\
        OFA-B & 78.1 & 397 & 5.1 &  &  \\
        OFA-L & 78.7 & 499 & 5.5 &  &  \\
        \rowcolor{LightOrange}
        OFA-T + Ours & 75.9 & 200 & 3.8 &  &  \\
        \rowcolor{LightOrange}
        OFA-S + Ours & 77.6 & 297 & 4.1 &  &  \\
        \rowcolor{LightOrange}
        OFA-B + Ours & 78.5 & 391 & 5.1 &  &  \\
        \rowcolor{LightOrange}
        OFA-L + Ours & 79.0 & 498 & 5.4 & \multirow{-4}{*}{\begin{tabular}[c]{@{}c@{}}0.09\\{\scriptsize\textcolor{LightRed}{↓4.7$\times$}}\end{tabular}} & \multirow{-4}{*}{\begin{tabular}[c]{@{}c@{}}250\\{\scriptsize\textcolor{LightRed}{↓4$\times$}}\end{tabular}} \\
        \midrule
        \multicolumn{6}{c}{ShuffleNet Search Space} \\
        \midrule
        SPOS & 73.7 & 323 & 3.5 & 0.32 & 1000 \\
        \rowcolor{LightOrange}
        SPOS + Ours & 74.4 & 325 & 3.7 & \begin{tabular}[c]{@{}c@{}}0.07\\{\scriptsize\textcolor{LightRed}{↓4.6$\times$}}\end{tabular} & \begin{tabular}[c]{@{}c@{}}250\\{\scriptsize\textcolor{LightRed}{↓4$\times$}}\end{tabular} \\
        \midrule
        \multicolumn{6}{c}{AutoFormer Search Space} \\
        \midrule
        AutoFormer-T & 74.7 & 1344 & 5.9 & \multirow{3}{*}{1.0} & \multirow{3}{*}{1000} \\
        AutoFormer-S & 81.6 & 4887 & 22.8 &  &  \\
        AutoFormer-B & 82.1 & 11305 & 54.0 &  &  \\
        \rowcolor{LightOrange}
        AutoFormer-T + Ours & 75.3 & 1366 & 6.0 &  &  \\
        \rowcolor{LightOrange}
        AutoFormer-S + Ours & 81.7 & 4897 & 22.9 &  &  \\
        \rowcolor{LightOrange}
        AutoFormer-B + Ours & 82.3 & 11074 & 52.8 & \multirow{-3}{*}{\begin{tabular}[c]{@{}c@{}}0.1\\{\scriptsize\textcolor{LightRed}{↓10$\times$}}\end{tabular}} & \multirow{-3}{*}{\begin{tabular}[c]{@{}c@{}}250\\{\scriptsize\textcolor{LightRed}{↓4$\times$}}\end{tabular}} \\
        \bottomrule
    \end{tabular}
\end{table}

In each macro search space, we adopt a two-stage NAS approach (\textit{i.e.,} OFA \cite{cai2020onceforall}, SPOS \cite{guo2020single}, and AutoFormer \cite{AutoFormer}) as the baseline and apply our collaborative LLM-based search to them. To ensure fair comparison, we directly reuse the pre-trained supernet from each baseline for the search phase. All baselines employ evolutionary algorithms as their search methods. Our approach implements Qwen3-30B-A3B \cite{qwen3technicalreport} as the foundational LLM, deployed locally via vLLM \cite{kwon2023efficient}, with temperature 0.6 and chain-of-thought reasoning enabled. 
Key experimental settings are summarized in \cref{tab:settings}, including evaluation methodologies, resource constraint types, and retraining requirements. 
Notably, while OFA originally employs a trained accuracy predictor for rapid subnet performance estimation, we observe that this predictor is unreliable for distinguishing top-performing architectures (Appendix D), and training the predictor requires substantial additional computational overhead. To ensure fair comparison, both OFA and \ours~adopt the same evaluation methodology as SPOS and AutoFormer by directly evaluating subnets on the full validation set, which improves OFA's accuracy over its original predictor.

All experiments are conducted on ImageNet \cite{deng2009imagenet} under varying resource constraints and fixed budgets on the number of explored architectures. Search costs, covering the entire duration of the search phase (including LLM inference), are quantified in GPU days on a single NVIDIA A100-80GB GPU.
For SPOS, we note the narrow FLOPs range (280-360M) of ShuffleNet search space is unsuitable for multi-tiered constraint decomposition. We therefore preserve the original setting from SPOS \cite{guo2020single}, searching exclusively at the 330M FLOPs constraint. Results report the best-performing subnet from three independent runs per method. More details about the search spaces are provided in Appendix A.

\paragraph{Micro Search Space.}
NAS-Bench-201 is a widely used cell-based search space for NAS benchmarking. It contains 15,625 architectures with precomputed performance on CIFAR-10, CIFAR-100, and ImageNet-16-120 datasets. Each architecture is represented as a directed acyclic graph (DAG) with 4 nodes and 6 edges, where every edge is assigned one of five candidate operations: \emph{none}, \emph{skip\_connect}, \emph{conv\_1x1}, \emph{conv\_3x3}, or \emph{avg\_pool\_3x3}.

We evaluate various approaches on NAS-Bench-201, including traditional one-shot NAS methods, emerging LLM-based NAS methods, and conventional search algorithms. 
To ensure fair comparison and avoid evaluation discrepancies, our method follows conventional search algorithms, \textit{i.e.,} Random Search (RS), Reinforcement Learning (RL), and Evolutionary Algorithm (EA), in directly adopting ground-truth performance metrics as accuracy measures. Specifically, EA employs a population size of 10 over 20 iterations, with 50\% elite preservation. RL operates with a learning rate of 0.01 and EMA momentum of 0.9. Our method explores up to 100 architectures. Results are averaged over 10 independent runs.

\subsection{Main Results}
\label{sec:experiment:main}
\paragraph{Macro Search Spaces Results.}

\cref{tab:macro-search-space} compares Top-1 test accuracy of subnets discovered by our method with baselines in different macro search spaces. 
Our method consistently outperforms baselines across all search spaces, with significantly lower search costs and fewer architecture evaluations, demonstrating its effectiveness and efficiency in discovering high-performing architectures. While our approach incurs additional overhead from LLM inference, this cost is substantially outweighed by the dramatic reduction in evaluations.
Specifically, our approach achieves up to 0.7\% accuracy improvements while reducing search costs by 4--10$\times$ compared to baselines.

\paragraph{Comparison with State-of-the-Art Methods.}

\begin{table}[!t]
    \caption{Comparison with SOTA NAS methods on ImageNet. Top: manual design. Upper middle: differentiable NAS. Lower middle: two-stage NAS. Bottom: LLM-based NAS.}
    \label{tab:comparison-with-sota}
    \small
    \centering
    \begin{tabular}{l|c|c|c}
        \toprule
        \multicolumn{1}{l|}{Method} & 
        \begin{tabular}[c]{@{}c@{}}Top-1 (\%)\end{tabular} & 
        \begin{tabular}[c]{@{}c@{}}Top-5 (\%)\end{tabular} & 
        \begin{tabular}[c]{@{}c@{}}FLOPs (M)\end{tabular} \\
        \midrule
        Mobilenetv2 \cite{sandler2018mobilenetv2} & 72.0 & - & 300 \\
        Mobilenetv3 \cite{howard2019searching} & 76.6 & - & 350 \\
        MixNet \cite{tan2019mixconv} & 77.0 & 93.3 & 360 \\
        \midrule
        DARTS \cite{liu2018darts} & 73.3 & 91.3 & 574 \\
        PC-DARTS \cite{xu2020pcdarts} & 75.8 & 92.7 & 597 \\
        DARTS- \cite{chu2021darts} & 76.2 & 93.0 & 467 \\
        EG-NAS \cite{cai2024eg} & 74.4 & - & - \\
        \midrule
        ProxylessNAS \cite{cai2018proxylessnas} & 74.6 & 92.2 & 320 \\
        SPOS \cite{guo2020single} & 75.6 & 92.8 & 330 \\
        FairNAS \cite{chu2021fairnas} & 76.7 & - & 325 \\
        OFA \cite{cai2020onceforall} & 77.5 & \underline{93.5} & 330 \\
        GreedyNAS \cite{you2020greedynas} & 76.8 & 93.0 & 324 \\
        GreedyNASv2 \cite{huang2022greedynasv2} & 77.5 & \underline{93.5} & 324 \\
        SUMNAS \cite{ha2022sumnas} & \underline{77.6} & - & 349 \\
        PA\&DA \cite{lu2023pa} & 77.3 & \underline{93.5} & 399 \\
        SparseNAS \cite{huang2024accelerating} & 76.7 & - & 295 \\
        DYNAS \cite{jeon2025subnet} & 75.9 & - & 329 \\
        \midrule
        GENIUS \cite{zheng2023can} & 74.9 & - & - \\
        RZ-NAS \cite{ji2025rznas} & 75.5 & - & - \\
        LM-Searcher \cite{hu2025lm} & 75.1 & - & - \\
        \rowcolor{LightOrange}
        Ours & \textbf{77.9} & \textbf{93.8} & 320 \\
        \bottomrule
    \end{tabular}
\end{table}

We further provide a comparison with SOTA NAS methods on ImageNet. As shown in \cref{tab:comparison-with-sota}, our approach achieves 77.9\% Top-1 accuracy with only 320M FLOPs, surpassing existing SOTA methods. This demonstrates our method's enhanced capability to efficiently navigate expansive search spaces and identify superior subnets. 

\paragraph{Micro Search Space Results.}
\begin{table*}[!t]
    \caption{Test and validation accuracy on NAS-Bench-201. Top: one-shot NAS methods. Upper middle: conventional search algorithms. Lower middle: LLM-based NAS methods. $^\dagger$ denotes results from \cite{ji2025rznas}.}
    \label{tab:nas-bench-201}
    \small
    \centering
    \begin{tabular}{l|cc|cc|cc}
        \toprule
        \multirow{2}{*}[-2pt]{Method} & \multicolumn{2}{c|}{CIFAR-10} & \multicolumn{2}{c|}{CIFAR-100} & \multicolumn{2}{c}{ImageNet-16-120} \\
        \cmidrule(lr){2-3} \cmidrule(lr){4-5} \cmidrule(lr){6-7}
        & valid & test & valid & test & valid & test \\
        \midrule
        DARTS \cite{liu2018darts} & 39.77 $\pm$ 0.00 & 54.30 $\pm$ 0.00 & 15.03 $\pm$ 0.00 & 15.61 $\pm$ 0.00 & 16.43 $\pm$ 0.00 &  16.32 $\pm$ 0.00 \\
        PC-DARTS \cite{xu2020pcdarts} & 89.96 $\pm$ 0.15 & 93.41 $\pm$ 0.30 & 67.12 $\pm$ 0.39 & 67.48 $\pm$ 0.89 & 40.83 $\pm$ 0.08 & 41.31 $\pm$ 0.22 \\
        DARTS- \cite{chu2021darts} & 91.03 $\pm$ 0.44 & 93.80 $\pm$ 0.40 & 71.36 $\pm$ 1.51 & 71.53 $\pm$ 1.51 & 44.87 $\pm$ 1.46 & 45.12 $\pm$ 0.82 \\
        FairNAS \cite{chu2021fairnas} & 90.07 $\pm$ 0.57 & 93.23 $\pm$ 0.18 & 70.94 $\pm$ 0.94 & 71.00 $\pm$ 1.46 & 41.90 $\pm$ 1.00 & 42.19 $\pm$ 0.31 \\
        EG-NAS \cite{cai2024eg} & 90.12 $\pm$ 0.05 & 93.56 $\pm$ 0.02 & 70.78 $\pm$ 0.12 & 70.91 $\pm$ 0.07 & 44.89 $\pm$ 0.29 & 46.13 $\pm$ 0.46 \\
        \midrule
        Random Search \cite{bergstra2012random} & 90.96 $\pm$ 0.24 & 93.83 $\pm$ 0.09 & 71.62 $\pm$ 0.73 & 71.52 $\pm$ 0.93 & 45.52 $\pm$ 0.48 & 45.37 $\pm$ 0.54 \\
        Reinforcement Learning \cite{williams1992simple} & 91.20 $\pm$ 0.41 & 93.94 $\pm$ 0.25 & 71.51 $\pm$ 0.93 & 71.78 $\pm$ 1.20 & 45.22 $\pm$ 0.79 & 45.95 $\pm$ 0.76 \\
        Evolutionary Algorithm \cite{real2019regularized} & 91.33 $\pm$ 0.35 & 94.23 $\pm$ 0.25 & 72.31 $\pm$ 1.07 & 72.82 $\pm$ 0.87 & 46.19 $\pm$ 0.46 & 46.49 $\pm$ 0.60 \\
        \midrule
        GENIUS$^\dagger$ \cite{zheng2023can} & 91.07 $\pm$ 0.20 & 93.79 $\pm$ 0.09 & 70.96 $\pm$ 0.33 & 70.91 $\pm$ 0.72 & 45.29 $\pm$ 0.81 & 44.96 $\pm$ 1.02 \\
        LLMatic$^\dagger$ \cite{nasir2024llmatic} & 91.42 $\pm$ 0.13 & \underline{94.26 $\pm$ 0.10} & 71.41 $\pm$ 1.44 & 71.62 $\pm$ 1.73 & 44.98 $\pm$ 0.87 & 45.87 $\pm$ 0.96 \\
        RZ-NAS$^\dagger$ \cite{ji2025rznas} & 91.45 $\pm$ 0.10 & 94.24 $\pm$ 0.12 & \underline{73.35 $\pm$ 0.14} & \underline{73.30 $\pm$ 0.21} & \underline{46.53 $\pm$ 0.24} & 46.24 $\pm$ 0.23 \\
        LM-Searcher \cite{hu2025lm} & \underline{91.52} & 94.20 & 72.82 & 72.96 & 46.48 & \underline{46.51}\\
        \rowcolor{LightOrange}
        Ours & \textbf{91.59 $\pm$ 0.04} & \textbf{94.37 $\pm$ 0.01} & \textbf{73.44 $\pm$ 0.12} & \textbf{73.44 $\pm$ 0.15} & \textbf{46.62 $\pm$ 0.10} & \textbf{46.79 $\pm$ 0.28} \\
        \midrule
        Optimal &  91.61 & 94.37 & 73.49 & 73.51 & 46.73 & 47.31 \\
        \bottomrule
    \end{tabular}
\end{table*}

We present NAS-Bench-201 results in \cref{tab:nas-bench-201}, where our approach performs favorably against other search methods across all datasets, especially achieving significant improvements over RL and EA. Our method further demonstrates enhanced robustness, evidenced by lower standard deviations.
Moreover, by exploring at most 100 architectures, significantly fewer than the other NAS methods require, our method achieves SOTA performance with high search efficiency.

\subsection{Discussion}
\label{sec:experiment:discussion}

\begin{figure}[!t]
    \centering
    \includegraphics[width=1\linewidth]{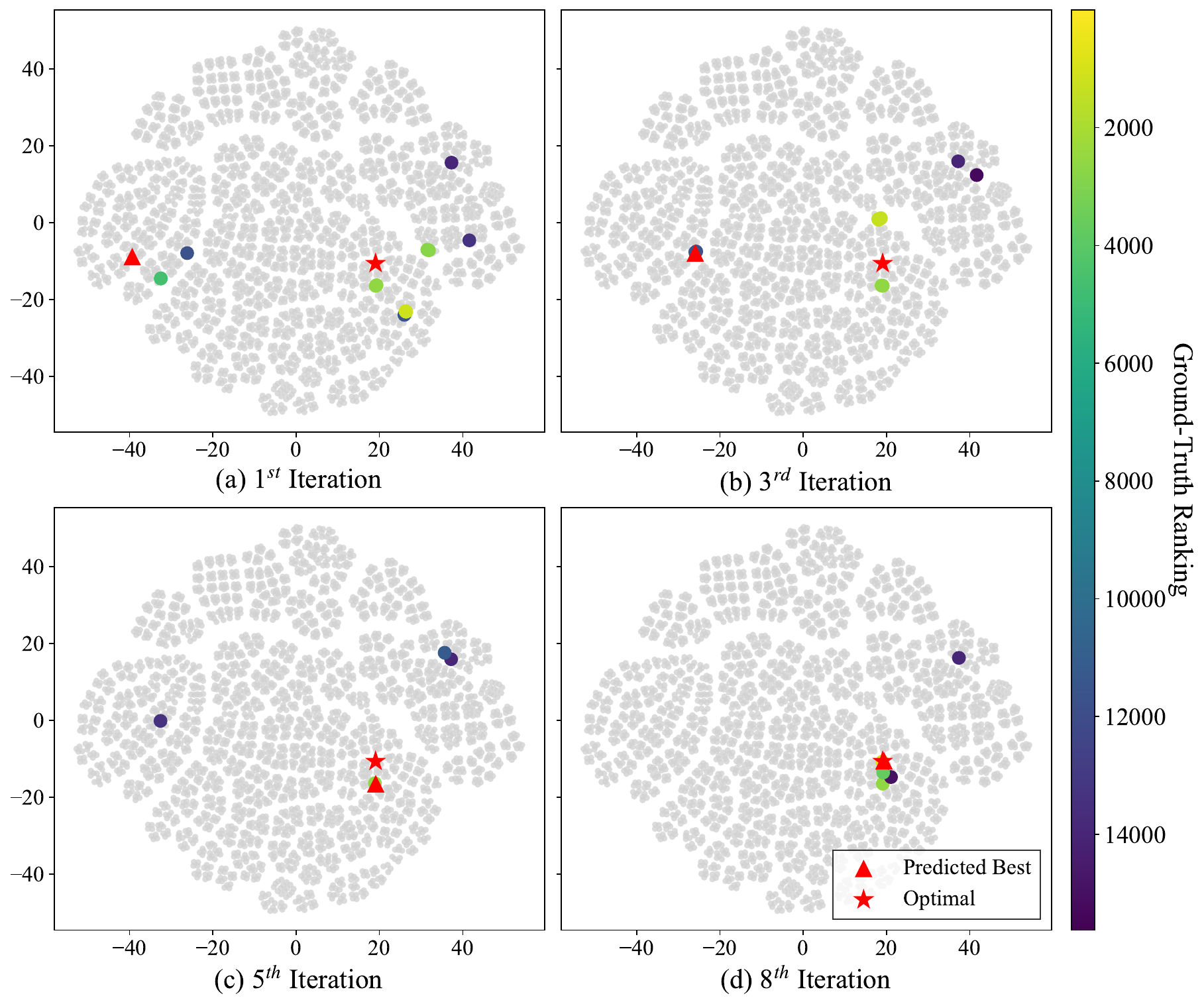}
    \caption{T-SNE visualization of \ours's search dynamics on ImageNet-16-120 within NAS-Bench-201 search space. Architectural performance rankings are represented through color-coding, with unexplored ones displayed in gray.}
    \label{fig:tsne}
\end{figure}

Traditional two-stage NAS relies on evolutionary algorithms (EAs) that refine architectures through stochastic operators like mutation and crossover. While capable of exploration, these operators are inherently local and undirected, generating new candidates via small perturbations of elite individuals. This approach lacks a global understanding of the performance landscape, often requiring numerous evaluations to achieve significant progress, resulting in reduced search efficiency.

In contrast, \ours~transforms architecture search into a directed, learning-based optimization process, aligning with the emerging \textit{"LLMs as optimizers"} paradigm \cite{yang2024large}, where LLMs can gradually improve the generated solutions based on past optimization steps. However, compared to prior work like OPRO \cite{yang2024large} that typically employs a single LLM to directly map optimization trajectories to solutions, our framework implements a more sophisticated, two-step generative process:
\begin{equation}
    \label{eq:CoLLM-NAS-search}
    \begin{aligned}
    \mathcal{S}_{t} &\gets \textsc{NavigatorLLM}(\mathcal{H}_{t}), \\
    \mathcal{C}_{t+1} &\gets \textsc{GeneratorLLM}(\mathcal{S}_{t}).
    \end{aligned}
\end{equation}
The \textit{Navigator LLM} first maps the optimization trajectory $\mathcal{H}_{t}$ to an abstract, natural-language strategy $\mathcal{S}_{t}$. Then, a separate \textit{Generator LLM} translates this strategy into concrete candidates $\mathcal{C}_{t+1}$. This "trajectory $\rightarrow$ strategy $\rightarrow$ solution" pipeline encourages more structured exploration by reasoning at a higher abstraction level, mitigating overfitting to specific architectural syntax and improving search robustness. 
We validate this hypothesis in \cref{sec:experiment:ablation:main-mechanisms}.

This guided search is further enhanced by the interplay between dual knowledge sources. First, as demonstrated in \cref{sec:method:arch-comprehension-in-llms}, LLMs' inherent knowledge of effective architecture design improves the quality of generated candidates and provides a powerful "warm-start" in promising regions.
Second, the progressive knowledge accumulated from optimization trajectories enables the \textit{Navigator LLM} to gradually learn an implicit model of the performance landscape, and guide the \textit{Generator LLM} towards increasingly promising regions. 
Moreover, our unique memory retention mechanism—featuring a stateful \textit{Navigator LLM} paired with a stateless \textit{Generator LLM}—further ensures a better balance between exploration and exploitation.
In \cref{fig:tsne}, we present t-SNE visualization of \ours's search dynamics, illustrating how the search distribution rapidly focuses on high-performance areas and efficiently locates the global optimum.

\subsection{Ablation Studies}
\label{sec:experiment:ablation}

\begin{figure}[!t]
    \centering
    \includegraphics[width=0.97\linewidth]{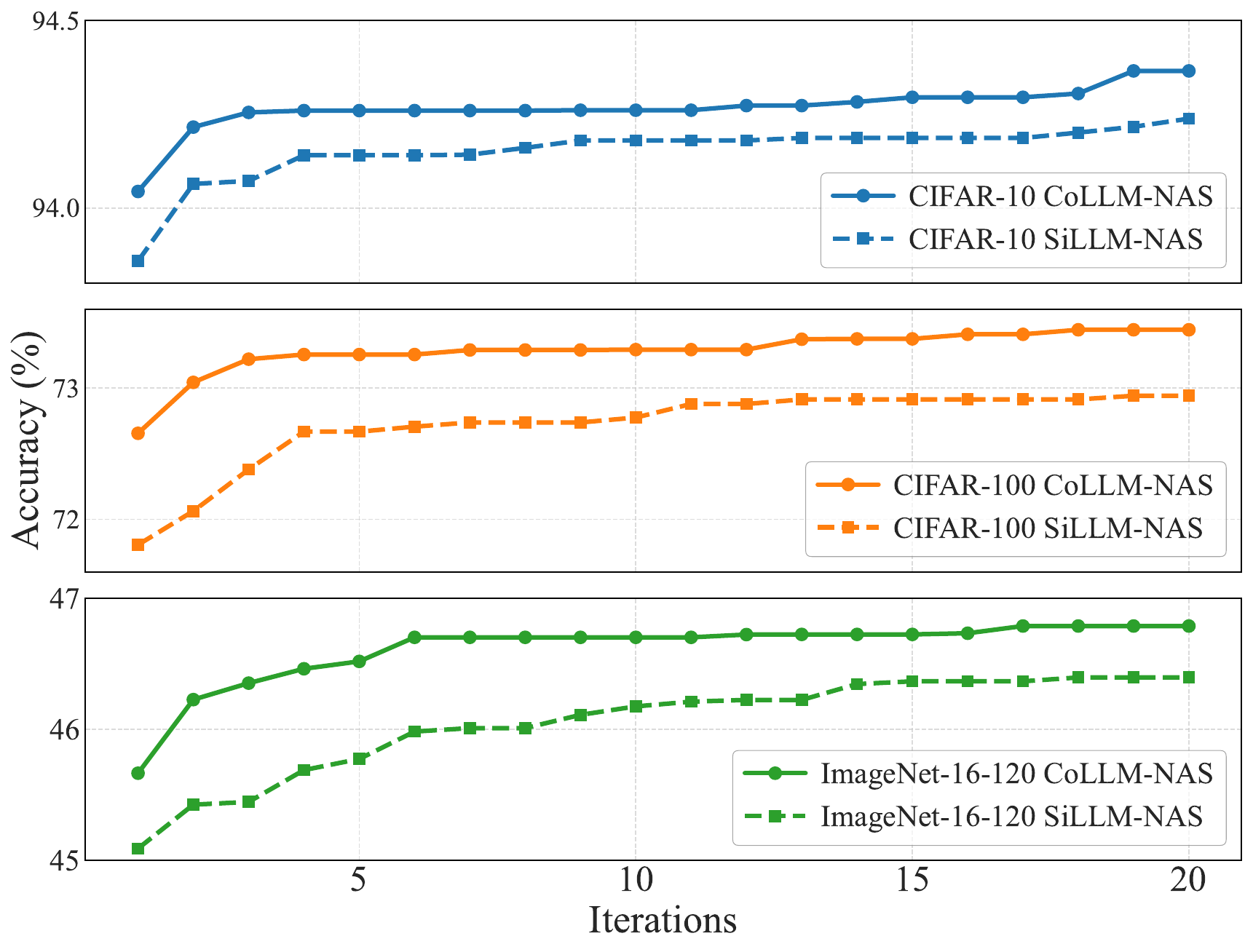}
    \caption{Comparison of iterative performance between \ours~and SiLLM-NAS on NAS-Bench-201.}
    \label{fig:ablation_collaboration}
\end{figure}

\paragraph{Main Mechanisms.}
\label{sec:experiment:ablation:main-mechanisms}
We ablate our collaboration and memory retention mechanisms respectively. For collaboration, we introduce a Single LLM NAS (SiLLM-NAS) variant that maintains the reflection-then-generation paradigm but consolidates both roles into a single LLM.
Over 10 runs, we track mean accuracy of the best architecture per iteration. \cref{fig:ablation_collaboration} shows \ours~consistently outperforms SiLLM-NAS across all datasets, verifying collaboration efficacy. Crucially, \ours~generates better initial populations, highlighting the \textit{Navigator LLM}'s critical role in initial exploration.

\begin{figure}[!t]
    \centering
    \includegraphics[width=0.97\linewidth]{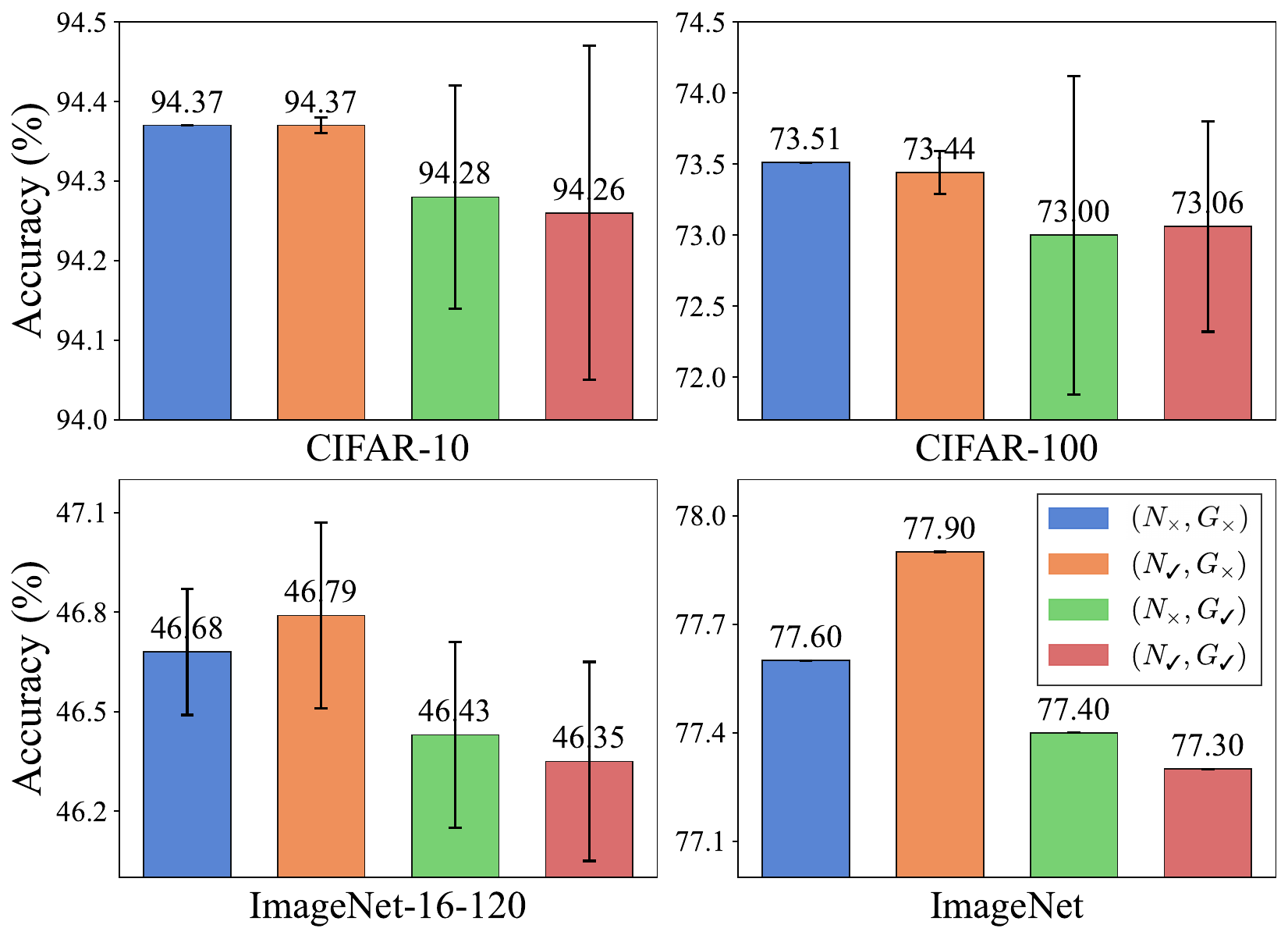}
    \caption{Impact of memory retention settings on different datasets. N\textsubscript{$\checkmark$/$\times$} and G\textsubscript{$\checkmark$/$\times$} denote whether the memory of \textit{Navigator}/\textit{Generator LLM} is retained.}
    \label{fig:ablation_memory}
\end{figure}

Regarding memory retention, \cref{fig:ablation_memory} shows that for low-complexity datasets (\textit{e.g.,} CIFAR-10, CIFAR-100), optimal performance is achieved without memory retention in either LLM, indicating that iterative feedback alone suffices for simpler tasks. Conversely, for high-complexity datasets (\textit{e.g.,} ImageNet-16-120, ImageNet), preserving the \textit{Navigator LLM}'s memory while disabling the \textit{Generator}'s yields optimal results, confirming the necessity of historical trajectory in demanding scenarios. 
Notably, we observe that retaining the \textit{Generator LLM}'s memory induces progressive noise accumulation, leading to performance degradation.

\paragraph{Rephrasing Prompts.} 
\label{sec:experiment:ablation:rephrasing-prompts}
To assess whether our gains depend on a particular wording, we rephrase the original prompts with three leading LLMs, \textit{i.e.,} Claude Sonnet 4 \cite{anthropic2025claudesonnet}, GPT-5 \cite{openai2025gpt5}, and DeepSeek-R1 \cite{guo2025deepseek}. As shown in \cref{tab:ablation_prompt}, all variants achieve comparable performance across NAS-Bench-201 datasets, with Variant 2 even outperforming the original prompt on ImageNet-16-120. The narrow performance spread indicates that improvements stem from the proposed collaborative framework rather than  handcrafted phrasing, evidencing robustness to linguistic reformulation.

\begin{figure}[!t]
    \centering
    \includegraphics[width=0.97\linewidth]{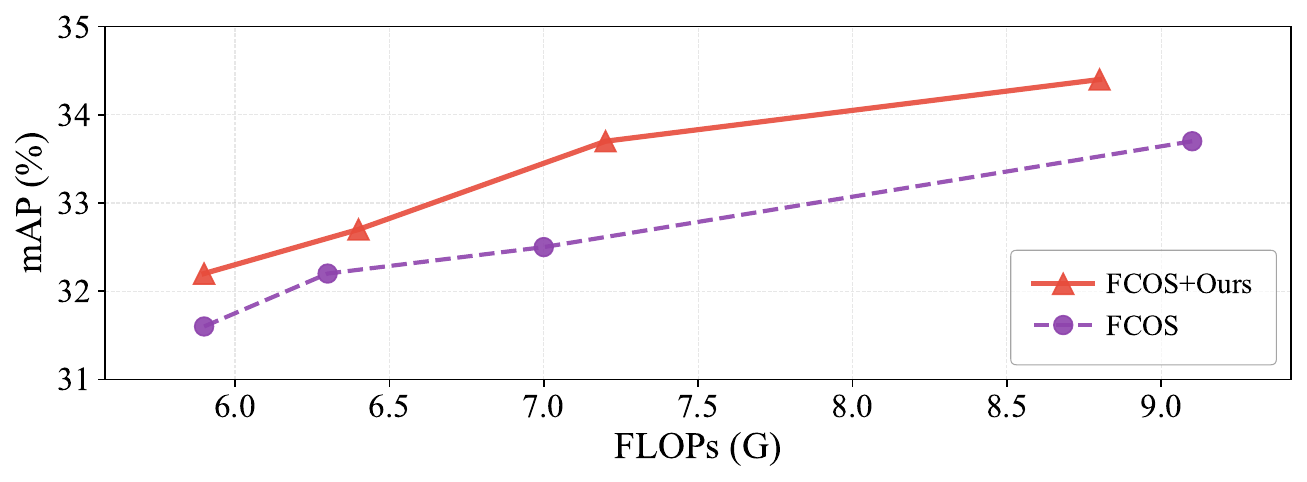}
    \caption{Transfer to FCOS object detection on COCO. FLOPs for backbone only.}
    \label{fig:object-detection}
\end{figure}

\paragraph{Different LLMs.}
\label{sec:experiment:ablation:different-llms}
We further perform extensive experiments with different open-source LLMs on NAS-Bench-201. As shown in \cref{tab:ablation_llm}, \ours~maintains consistently strong performance across diverse LLMs, confirming its generality beyond specific LLM implementations.
Additionally, we verify \ours's robustness across different LLM temperature settings. Detailed experiments and results are provided in Appendix B.

\begin{table}[!t]
    \caption{Performance comparison with rephrasing prompts. Variant 1-3 are rephrased by Claude Sonnet 4, GPT-5, and DeepSeek-R1 respectively.}
    \label{tab:ablation_prompt}
    \small
    \setlength{\tabcolsep}{1.8mm}
    \centering
    \begin{tabular}{l|ccc}
        \toprule
        Prompt & CIFAR-10 & CIFAR-100 & ImageNet-16-120 \\
        \midrule
        Base (Ours) & \textbf{94.37$\pm$0.01} & \textbf{73.44$\pm$0.15} & \underline{46.79$\pm$0.28} \\
        Variant 1 & \underline{94.36$\pm$0.02} & 73.35$\pm$0.52 & 46.52$\pm$0.36 \\
        Variant 2 & 94.35$\pm$0.03 & \underline{73.36$\pm$0.18} & \textbf{46.89$\pm$0.35} \\
        Variant 3 & 94.16$\pm$0.23 & 73.19$\pm$0.11 & 46.60$\pm$0.29 \\
        \bottomrule
    \end{tabular}
\end{table}

\begin{table}[!t]
    \caption{Performance comparison using different LLMs. $^*$ from \cite{qwen3technicalreport}, and $^\ddagger$ from \cite{guo2025deepseek}.}
    \label{tab:ablation_llm}
    \small
    \setlength{\tabcolsep}{1.5mm}
    \centering
    \begin{tabular}{l|ccc}
        \toprule
        \multirow{2}{*}{LLM} & \multirow{2}{*}{CIFAR-10} & \multirow{2}{*}{CIFAR-100} & ImageNet- \\
        & & & 16-120 \\
        \midrule
        Qwen3-30B-A3B$^*$ & \underline{94.37$\pm$0.01} & \textbf{73.44$\pm$0.15} & \textbf{46.79$\pm$0.28} \\
        \midrule
        Qwen3-32B$^*$ & 94.31$\pm$0.14 & 73.29$\pm$0.29 & 46.64$\pm$0.52 \\
        \midrule
        DeepSeek-R1- & \multirow{2}{*}{94.36$\pm$0.04} & \multirow{2}{*}{73.37$\pm$0.19} & \multirow{2}{*}{46.53$\pm$0.32} \\
        Distill-Qwen-32B$^\ddagger$ & & & \\
        \midrule
        DeepSeek-R1- & \multirow{2}{*}{\textbf{94.37$\pm$0.00}} & \multirow{2}{*}{\underline{73.41$\pm$0.22}} & \multirow{2}{*}{\underline{46.74$\pm$0.31}} \\
        Distill-Llama-70B$^\ddagger$ & & & \\
        \bottomrule
    \end{tabular}
\end{table}


\subsection{Downstream Tasks}
\label{sec:experiment:downstream-tasks}

We transfer architectures discovered by \ours~to object detection tasks. Specifically, we adopt architectures from MobileNet search space as the backbone of FCOS detector \cite{tian2020fcos}, inheriting pre-trained supernet weights and training on COCO dataset for a 1x schedule. \cref{fig:object-detection} shows that our architectures perform favorably, demonstrating strong generalization to downstream tasks.

%% file: sections/05_conclusion.tex
\section{Conclusion}
\label{sec:conclusion}
In this paper, we present \ours, a collaborative LLM-based NAS framework that integrates LLMs with two-stage NAS. We design a stateful \textit{Navigator LLM} to provide adaptive search strategies, a stateless \textit{Generator LLM} to synthesize high-quality architectures, and a \textit{Coordinator} module to orchestrate inter-LLM communication and manage evaluation processes.
\ours~employs knowledge-guided search by coupling LLMs' inherent knowledge of structured neural architectures with progressive knowledge from iterative feedback and historical trajectory.
Extensive experiments demonstrate that our approach consistently enhances various two-stage NAS methods across diverse search spaces, outperforms existing NAS methods and conventional search algorithms, and achieves new SOTA results while reducing search costs significantly.
Furthermore, the generalizable nature of \ours~suggests promising extensions beyond NAS, revealing avenues for future exploration.

%% file: sections/06_acknowledgement.tex
\section{Acknowledgments}
\label{sec:acknowledgement}
This work was supported by National Natural Science Foundation of China under Grant 62176007.

%% file: appendix/00_macro_search_spaces.tex
\section{Macro Search Spaces}
\label{app:macro_search_spaces}

\paragraph{MobileNet Search Space.} 
The MobileNet search space employs MBConv blocks as fundamental units, featuring depthwise separable convolutions and squeeze-excitation modules. Configurable dimensions include: 
\begin{itemize}
\item \textit{Resolution} ($r \in \{160, 176, 192, 208, 224\}$) controlling input scale, 
\item \textit{Stage depth} ($d_i \in \{2,3,4\}$ for $i=1$ to $5$ stages) determining active blocks per stage, 
\item \textit{Kernel size} ($k_j \in \{3,5,7\}$ for $j=1$ to $20$ convolutional layers), 
\item \textit{Expansion ratio} ($e_j \in \{3,4,6\}$ for $j=1$ to $20$ inverted residual blocks).
\end{itemize} 
The combinatorial space contains $\sim$$10^{19}$ architectures, with stage-specific depth controlling block activation patterns (\textit{e.g.,} $d=[2,3,4,3,2]$ activates blocks 0-1,4-6,8-11,12-14,16-17). 

\paragraph{ShuffleNet Search Space.} 
The ShuffleNet search space utilizes ShuffleNetv2 units and Xception modules as fundamental building blocks, featuring channel split/shuffle operations and depthwise separable convolutions. Configurable dimensions include:
\begin{itemize}
\item ShuffleNet unit with $3\times3$ kernel, 
\item ShuffleNet unit with $5\times5$ kernel, 
\item ShuffleNet unit with $7\times7$ kernel, 
\item Xception module. 
\end{itemize}
This yields $4^{20}$ possible configurations, where each block's operator is selected independently during sampling.

\paragraph{AutoFormer Search Space.}
The AutoFormer search space employs multi-head self-attention (MSA) and MLP blocks as fundamental units, defining a pure-transformer search space with four layer-wise variables: 
\begin{itemize}
\item \textit{Depth}: Tiny $\in \{12, 13, 14\}$, Small $\in \{12, 13, 14\}$, and Base $\in \{14, 15, 16\}$.
\item \textit{Embedding dimension}: Tiny $\in \{192, 216, 240\}$, Small $\in \{320, 384, 448\}$, and Base $\in \{528, 576, 624\}$.
\item \textit{Number of heads}: Tiny $\in \{3, 4\}$, Small $\in \{5, 6, 7\}$, and Base $\in \{8, 9, 10\}$.
\item \textit{MLP ratio} $\in \{3.0, 3.5, 4.0\}$ uniformly across all model scales.
\end{itemize}
The unified space exceeds $10^{16}$ configurations with layer-wise independent hyperparameters.

%% file: appendix/01_temperature.tex
\section{Ablation on Temperature Settings}
\label{app:ablation_temperature}
\begin{figure}[ht]
    \centering
    \includegraphics[width=0.4\linewidth]{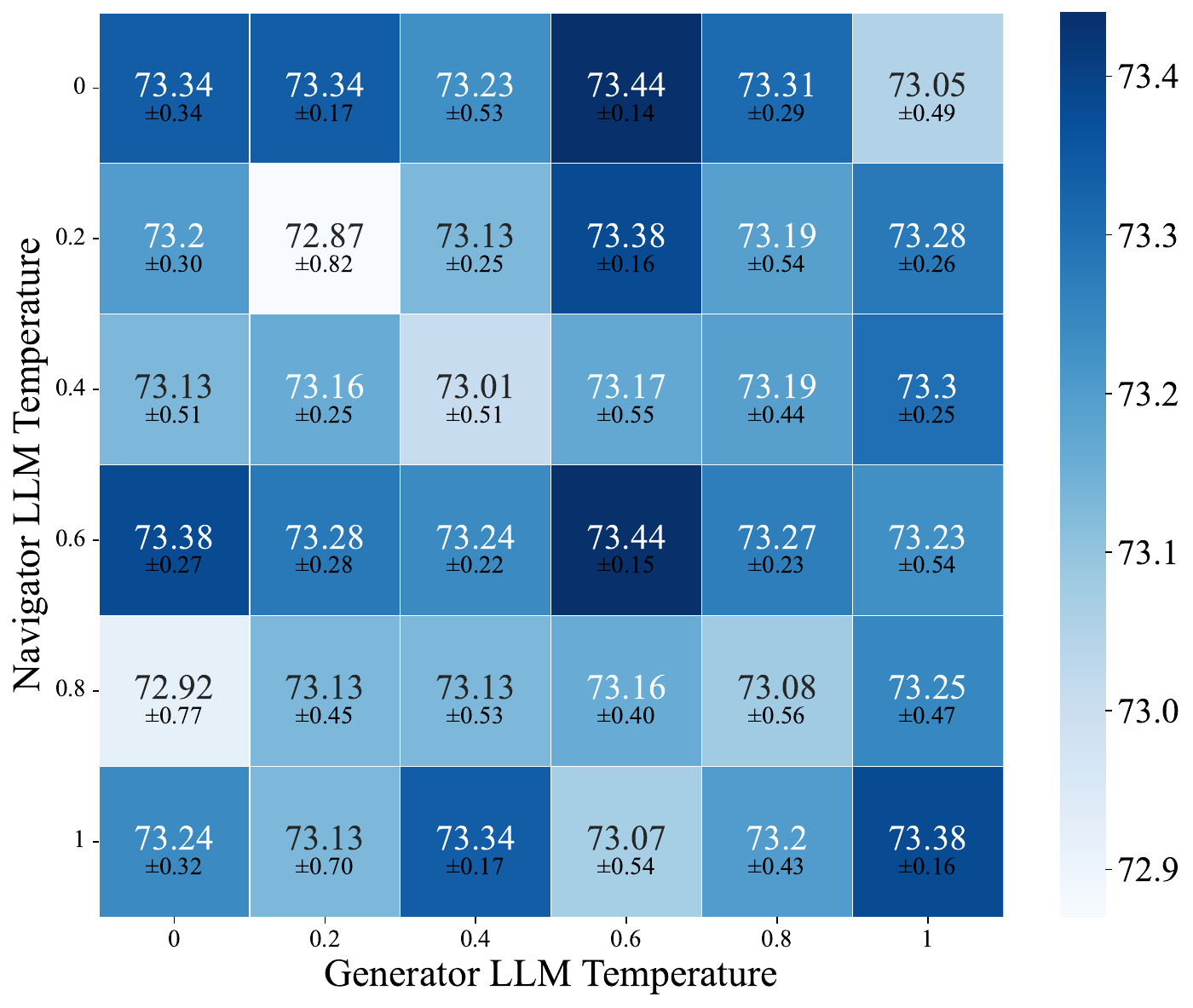}
    \caption{Performance comparison of different temperature settings on CIFAR-100 within NAS-Bench-201 search space.}
    \label{fig:temperature_ablation}
\end{figure}

To investigate the impact of LLM temperature on our approach, we conduct a comprehensive sensitivity analysis using six distinct temperature values $[0, 0.2, 0.4, 0.6, 0.8, 1.0]$ for both LLMs. This 6$\times$6 grid experiment, conducted on CIFAR-100 test set within NAS-Bench-201 search space, evaluates performance consistency across all temperature combinations. Robustness is quantified using the coefficient of variation (CV) of results. As shown in Figure~\ref{fig:temperature_ablation}, our method maintains consistently high performance across all temperature settings, with a remarkably low CV of 0.1769\%, demonstrating exceptional robustness to temperature variations. We adopt a temperature of 0.6 for both LLMs as the optimal setting, enabling each to independently achieve optimal performance.

%% file: appendix/02_PoC_experiment.tex
\section{Proof-of-Concept Experiment}
\label{app:poc}
We provide the sampled architectures from our proof-of-concept experiment and their performance metrics on CIFAR-10 test set, along with the prompts provided to the LLM and an example response.

\subsection{Sampled Architectures}
\label{app:poc:sampled}
\begin{table}[!ht]
\caption{Sampled architectures and their performance metrics on CIFAR-10 test dataset.}
\small
\centering
\begin{tabular}{c|p{10cm}|c|c}
\toprule
ID & Architecture & Top-1 (\%) & Ranking \\
\midrule
\multirow{1}{*}{1} & \texttt{$|$none$\sim$0$|$+$|$none$\sim$0$|$none$\sim$1$|$+$|$none$\sim$0$|$none$\sim$1$|$skip\_connect$\sim$2$|$} & \multirow{1}{*}{10.00} & \multirow{1}{*}{10} \\
\midrule
\multirow{2}{*}{2} & \texttt{$|$none$\sim$0$|$+$|$none$\sim$0$|$none$\sim$1$|$+$|$nor\_conv\_1x1$\sim$0$|$} & \multirow{2}{*}{88.67} & \multirow{2}{*}{7} \\
 & \texttt{$|$nor\_conv\_1x1$\sim$1$|$skip\_connect$\sim$2$|$} & & \\
\midrule
\multirow{2}{*}{3} & \texttt{$|$nor\_conv\_3x3$\sim$0$|$+$|$nor\_conv\_3x3$\sim$0$|$nor\_conv\_3x3$\sim$1$|$+} & \multirow{2}{*}{94.37} & \multirow{2}{*}{1} \\
 & \texttt{$|$skip\_connect$\sim$0$|$nor\_conv\_3x3$\sim$1$|$nor\_conv\_1x1$\sim$2$|$} & & \\
\midrule
\multirow{2}{*}{4} & \texttt{$|$nor\_conv\_3x3$\sim$0$|$+$|$skip\_connect$\sim$0$|$nor\_conv\_1x1$\sim$1$|$+} & \multirow{2}{*}{92.98} & \multirow{2}{*}{2} \\
 & \texttt{$|$nor\_conv\_3x3$\sim$0$|$nor\_conv\_1x1$\sim$1$|$nor\_conv\_3x3$\sim$2$|$} & & \\
\midrule
\multirow{2}{*}{5} & \texttt{$|$avg\_pool\_3x3$\sim$0$|$+$|$none$\sim$0$|$none$\sim$1$|$+$|$skip\_connect$\sim$0$|$} & \multirow{2}{*}{86.63} & \multirow{2}{*}{8} \\
 & \texttt{$|$none$\sim$1$|$none$\sim$2$|$} & & \\
\midrule
\multirow{2}{*}{6} & \texttt{$|$none$\sim$0$|$+$|$nor\_conv\_1x1$\sim$0$|$avg\_pool\_3x3$\sim$1$|$+} & \multirow{2}{*}{89.53} & \multirow{2}{*}{6} \\
 & \texttt{$|$nor\_conv\_1x1$\sim$0$|$nor\_conv\_3x3$\sim$1$|$nor\_conv\_1x1$\sim$2$|$} & & \\
\midrule
\multirow{2}{*}{7} & \texttt{$|$nor\_conv\_1x1$\sim$0$|$+$|$nor\_conv\_3x3$\sim$0$|$nor\_conv\_1x1$\sim$1$|$+} & \multirow{2}{*}{92.36} & \multirow{2}{*}{3} \\
 & \texttt{$|$nor\_conv\_1x1$\sim$0$|$skip\_connect$\sim$1$|$skip\_connect$\sim$2$|$} & & \\
\midrule
\multirow{2}{*}{8} & \texttt{$|$avg\_pool\_3x3$\sim$0$|$+$|$none$\sim$0$|$avg\_pool\_3x3$\sim$1$|$+} & \multirow{2}{*}{78.71} & \multirow{2}{*}{9} \\
 & \texttt{$|$skip\_connect$\sim$0$|$avg\_pool\_3x3$\sim$1$|$none$\sim$2$|$} & & \\
\midrule
\multirow{2}{*}{9} & \texttt{$|$nor\_conv\_3x3$\sim$0$|$+$|$none$\sim$0$|$avg\_pool\_3x3$\sim$1$|$+} & \multirow{2}{*}{92.03} & \multirow{2}{*}{4} \\
 & \texttt{$|$nor\_conv\_3x3$\sim$0$|$skip\_connect$\sim$1$|$avg\_pool\_3x3$\sim$2$|$} & & \\
\midrule
\multirow{2}{*}{10} & \texttt{$|$nor\_conv\_3x3$\sim$0$|$+$|$avg\_pool\_3x3$\sim$0$|$avg\_pool\_3x3$\sim$1$|$+} & \multirow{2}{*}{90.75} & \multirow{2}{*}{5} \\
 & \texttt{$|$nor\_conv\_3x3$\sim$0$|$none$\sim$1$|$skip\_connect$\sim$2$|$} & & \\
\bottomrule
\end{tabular}
\label{tab:poc}
\end{table}

\subsection{Prompts}
\label{app:poc:prompts}
\begin{codebox}
"You are the Architecture Ranking Expert, specializing in evaluating and ranking neural architectures for image classification tasks. You possess prior knowledge of effective architectural patterns and use this knowledge to assess the potential performance of different architectures. \\
\\
\# Core Mission \\
Your primary task is to rank a given set of neural architectures from highest to lowest expected performance on CIFAR-10 dataset. \\
\\
\# Architecture Knowledge Base \\
\#\# Overall architecture description \\
The entire network architecture is composed of the same cell stacked multiple times, as well as some fixed pre- and post-processing modules. Therefore, you only need to focus on the internal connections and operations of this cell, without considering the rest of the network. \\
\\
\#\# Search Space \\
The search space is defined as follows: \\
- Each cell contains 4 nodes (node 0-3) \\
- Connections between nodes are represented as a directed acyclic graph (DAG) \\
- Each edge can choose one of the following 5 operations: 

\quad* none: no connection

\quad* skip\_connect: skip connection

\quad* nor\_conv\_1x1: 1x1 convolution

\quad* nor\_conv\_3x3: 3x3 convolution

\quad* avg\_pool\_3x3: 3x3 average pooling \\
\newline
\#\# Representation Format \\
Valid architectures follow this representation format: \\ 
`|op1$\sim$0|+|op2$\sim$0|op3$\sim$1|+|op4$\sim$0|op5$\sim$1|op6$\sim$2|` \\
This format represents connections between 4 nodes (0-3) with ordered connections: \\
1.First section (`|op1$\sim$0|`): one connection from node 0 to node 1 \\
2.Second section (`|op2$\sim$0|op3$\sim$1|`): two connections to node 2:

\quad- First must be from node 0 (`op2$\sim$0`)

\quad- Second must be from node 1 (`op3$\sim$1`)

3.Third section (`|op4$\sim$0|op5$\sim$1|op6$\sim$2|`):three connections to node 3:

\quad- First must be from node 0 (`op4$\sim$0`)

\quad- Second must be from node 1 (`op5$\sim$1`)

\quad- Third must be from node 2 (`op6$\sim$2`) \\
Example: 
`|nor\_conv\_3x3$\sim$0|+|nor\_conv\_3x3$\sim$0|avg\_pool\_3x3$\sim$1|+|skip\_connect
$\sim$0|nor\_conv\_3x3$\sim$1|skip\_connect$\sim$2|`

\quad- Node 1: Apply 3x3 convolution to node 0 

\quad- Node 2: Apply 3x3 convolution to node 0, apply 3x3 average pooling to node 1, then add them 

\quad- Node 3: Apply skip connection to node 0, apply 3x3 convolution to node 1, apply skip connection to node 2, then add them. \\
\newline
Here are the architectures need to be ranked: \{archs\}. When outputting, you only need to output the predicted ranking and architecture number in descending order of predicted ranking."
\end{codebox}

\subsection{Response}
\label{app:poc:response}
\begin{responsebox}
"reasoning\_content":"

Okay, let's analyze these 10 architectures for CIFAR-10. The main idea is to look at each cell of 4 nodes and the operations on each edge: none, skip\_connect, nor\_conv\_1x1, nor\_conv\_3x3, and avg\_pool\_3x3. As we know, 3x3 convolutions are really good for capturing spatial features, skip connections help with gradient flow, and pooling or none operations tend to reduce the cell's representational power. \\
\newline
Starting with Architecture 1, most operations are none, except a skip at node 3. So honestly, it's very weak. Architecture 2 is slightly better because node 3 has two 1x1 convolutions and a skip, but nodes 1 and 2 contribute almost nothing. Architecture 5 and 8 rely heavily on avg\_pool or none operations, so they probably won't perform well either. \\
\newline
Now looking at Architecture 3, node 1 has a 3x3 convolution, node 2 combines two 3x3 convolutions, and node 3 has a skip, 3x3, and 1x1. This is quite solid---strong spatial features and diverse aggregation paths, which should really help. Architecture 4 is similar: node 1 has 3x3, node 2 has a skip and 1x1, and node 3 combines 3x3, 1x1, 3x3. It's also strong, though node 2 has fewer spatial convolutions compared to 3. Architecture 6 mixes 1x1, 3x3, and avg\_pool, but node 1 is none, which weakens early feature extraction. Architecture 7 has 1x1 at node 1, node 2 with 3x3 and 1x1, and node 3 with 1x1 and two skips---still decent, though the skips in node 3 don't add as much spatial power. Architecture 9 has an avg\_pool at node 2, reducing features, and node 3 combines 3x3, skip, and avg\_pool. Architecture 10 is similar but node 2 has two avg\_pools, which really limits intermediate representation, and node 3 combines 3x3, skip, and none. \\
\newline
So, if we think about it, architectures with more 3x3 convolutions and fewer none or pooling operations, along with skip connections in later nodes, are likely to perform better. Considering all this, the expected ranking from highest to lowest would be: Architecture 3 at the top, then Architecture 4, followed by Architecture 7, Architecture 6, Architecture 9, Architecture 10, Architecture 2, Architecture 5, Architecture 8, and finally Architecture 1." \\
\newline
"content":" \\
Ranking 1. Architecture 3 \\
Ranking 2. Architecture 4 \\
Ranking 3. Architecture 7 \\
Ranking 4. Architecture 6 \\
Ranking 5. Architecture 9 \\
Ranking 6. Architecture 10 \\
Ranking 7. Architecture 2 \\
Ranking 8. Architecture 5 \\
Ranking 9. Architecture 8 \\
Ranking 10. Architecture 1"
\end{responsebox}

%% file: appendix/03_ofa_acc_predictor.tex
\section{Analysis of Accuracy Predictor}
\label{app:ofa-accuracy-predictor}

The Accuracy Predictor in OFA is trained as follows: After obtaining the trained supernet, 16K subnets with different architectures and input resolutions are randomly sampled, and their accuracy is measured on 10K validation images. These [architecture, accuracy] pairs train a three-layer feedforward neural network predictor for rapid accuracy estimation.

To investigate the predictor's reliability, particularly for top-performing architectures, we conduct the following experiment: Under the 400M FLOPs constraint, we sample 15 elite architectures discovered by OFA, then rank them in descending order based on both predicted accuracy from the predictor and actual accuracy evaluated on the full validation set. 
The ranking consistency between these two orderings is quantified through Kendall's $\tau$.

Across three independent trials, Kendall's $\tau$ values are 0.410 (p=0.036), 0.448 (p=0.021), and 0.467 (p=0.016), yielding an average of \textbf{0.44}. This indicates only a statistically moderate correlation, demonstrating the limited reliability of the predictor to distinguish top-performing architectures. This limitation is directly related to the predictor's training methodology, which can only incorporate a subset of possible architectures and thus fails to adequately capture the distribution of top-performing candidates.

%% file: appendix/04_prompts.tex
\section{Prompts and Responses}
\label{app:prompts_and_responses}
Below, we present illustrative prompts and responses of \ours~within MobileNet search space. 

\subsection{\textit{Navigator LLM}}
\label{app:prompts_and_responses:navigator}

\subsubsection{System Prompt}
\label{app:prompts_and_responses:navigator:system}

\begin{itemize}
\item Role Definition
\end{itemize}
\begin{codebox}
"You are the Navigator\_LLM, an expert neural architecture analyst specializing in identifying patterns and improvement opportunities for neural architectures. You possess deep knowledge of neural architecture design principles and work collaboratively with a Generator\_LLM to efficiently discover high-performing architectures."
\end{codebox}

\begin{itemize}
\item Collaborative Responsibility
\end{itemize}
\begin{codebox}
"\# Your Role in the Collaboration \\
Your responsibility is to analyze the performance patterns of generated architectures and develop important insights to guide the Generator\_LLM toward more promising areas of the search space. The Generator\_LLM relies on your expertise to efficiently navigate the search space."
\end{codebox}

\begin{itemize}
\item Core Mission and Objectives
\end{itemize}
\begin{codebox}
"\# Core Mission \\
Your primary objective is to guide the search toward architectures that achieve $>$\{self.expected\_acc\}\% accuracy on ImageNet while satisfying the computational constraint of <\{self.max\_flops\}M FLOPs. Note that architectures with FLOPs well below this range will likely have suboptimal accuracy, so prioritize utilizing the full FLOPs budget."
\end{codebox}

\begin{itemize}
\item Knowledge about search space
\end{itemize}
\begin{codebox}
"\# Architecture Knowledge Base \\
\#\# Search Space \\
The search space is defined as follows: \\
When Generating Architectures \\
- Resolution (r): ONE value from [160, 176, 192, 208, 224] \\
- Depth (d): 5 values, each from [2, 3, 4] \\
- Kernel sizes (ks): 20 values, each from [3, 5, 7] \\
- Expansion ratios (e): 20 values, each from [3, 4, 6]

\#\# Representation Format \\
Architectures are represented in the format: \{'r': [r1], 'd': [d1,d2,d3,d4,d5], 'ks': [k1,k2,...,k20], 'e': [e1,e2,...,e20]\} \\
For example: \{'r': [176], 'd': [2, 3, 3, 4, 4], 'ks': [5, 3, 7, 3, 5, 5, 7, 3, 3, 5, 5, 3, 5, 5, 7, 7, 3, 7, 5, 3], 'e': [3, 3, 6, 3, 4, 4, 3, 6, 4, 6, 4, 4, 4, 4, 3, 3, 4, 6, 6, 3]\}

\#\# Structural Details \\
- The network has 5 stages, each containing up to 4 blocks (20 blocks total) \\
- The depth values determine which blocks are active in each stage, and only active blocks affect performance:

\quad- d[0] determines active blocks in stage 0 (blocks 0-3) 

\quad- d[1] determines active blocks in stage 1 (blocks 4-7) 

\quad- d[2] determines active blocks in stage 2 (blocks 8-11) 

\quad- d[3] determines active blocks in stage 3 (blocks 12-15) 

\quad- d[4] determines active blocks in stage 4 (blocks 16-19) 

- Example: d=[2,3,4,3,2] means blocks 0-1, 4-6, 8-11, 12-14, 16-17 are active"
\end{codebox}

\begin{itemize}
\item Collaboration Workflow
\end{itemize}
\begin{codebox}
"\# Collaboration Process \\
The search follows an iterative refinement loop: \\
\newline
**Iteration 0 (Initial Exploration)**: \\
- You provide an initial exploration strategy emphasizing architectural diversity \\
- Generator\_LLM creates a diverse population based on your strategy \\
- Coordinator evaluates all generated architectures and collects performance data \\
\newline
**Iteration N $\geq$ 1 (Iterative Refinement)**: \\
- You receive evaluation results from the previous iteration \\
- You analyze performance patterns and refine your strategy, progressively focusing on high-performing regions \\
- Generator\_LLM creates new architectures following your refined strategy \\
- Coordinator evaluates the new architectures and provides feedback"
\end{codebox}

\begin{itemize}
\item Output Format
\end{itemize}
\begin{codebox}
"\# When providing guidance: \\
Use this format to provide guidance: \\
``` \\
\# SUMMARY

[sentences summarizing key insights from evaluated architectures] \\

\# KEY PATTERNS

- [Pattern 1: Important statistical observation] \\
- [Pattern 2: Important statistical observation] \\
- [Pattern 3: Important statistical observation] \\
... \\
\newline
\# NAVIGATION DIRECTIONS

[Provide specific, actionable guidance for each architectural dimension:] \\
\#\# Resolution

[Guidance on which resolution values to use and their expected impact] \\
\#\# Depth

[Guidance on depth distribution across 5 stages, including specific configurations] \\
\#\# Kernel Size

[Guidance on kernel size selection for different stages/blocks] \\
\#\# Expansion Ratio

[Guidance on expansion ratio patterns across stages/blocks] \\

[Optional for later iterations: Additional sections like CRITICAL CONSTRAINTS, RECOMMENDED RANGES, or EXAMPLE ARCHITECTURES to provide more specific guidance] \\
```"
\end{codebox}

\subsubsection{User Prompt}
\label{app:prompts_and_responses:navigator:user}

\begin{itemize}
\item Initial Strategy (Iteration 0)
\end{itemize}
\begin{codebox}
"You are tasked with developing an initial exploration strategy for generating a diverse population of neural architectures. This is the first iteration, and we have no prior evaluation data. Your goal is to establish a broad exploration strategy that covers diverse promising regions of the search space. \\
\newline
Please provide an initial strategy that emphasizes architectural diversity while incorporating your knowledge of effective neural architecture design principles. Include specific guidance on: \\
- How to select resolution values to balance computational efficiency and representational capacity \\
- How to distribute depth values across the 5 stages to capture multi-level features \\
- What patterns to explore for kernel sizes and expansion ratios based on your understanding of convolutional architectures \\
- Any design principles or constraints that could help initialize the search in promising regions \\
\newline
Remember, the Generator\_LLM will rely on your strategy to create the initial population. Focus on broad exploration rather than narrow optimization at this stage."
\end{codebox}

\begin{itemize}
\item Refined Strategy (Iteration $N \geq 1$)
\end{itemize}
\begin{codebox}
"Here are the generated architectures and their performance guided by your previous strategy: \{evaluation\_results\} \\
\newline
Based on these results, please rethink and revise your strategy for the next generation of architectures. \\
Focus on: \\
- Patterns you observe in the better performing architectures \\
- Specific parameter ranges or combinations that should be explored more \\
- Clear guidance for generating the next set of architectures \\
- Any trends in how different parameters affect accuracy and FLOPs \\
\newline
So far, the best architecture has accuracy \{best\_accuracy\}\% with FLOPs constraint of <\{self.max\_flops\}M. Our target is to exceed \{self.expected\_acc\}\% accuracy while keeping FLOPs <\{self.max\_flops\}M."
\end{codebox}

\subsubsection{Responses}
\label{app:prompts_and_responses:navigator:responses}

\begin{itemize}
\item Initial Strategy (Iteration 0)
\end{itemize}
\begin{responsebox}
"reasoning\_content":... \\
"content":" \\
\# SUMMARY \\
For the initial exploration, we will generate a diverse population covering broad regions of the search space. The strategy emphasizes architectural diversity across all dimensions (resolution, depth, kernel size, expansion ratio) while incorporating fundamental design principles. We will explore various resolution-depth trade-offs, different kernel size distributions, and varied expansion ratio patterns to identify promising architectural characteristics. \\
\newline
\# KEY PATTERNS \\
- **Resolution and Depth Trade-off**: Higher resolutions (224) may require lower depth values to stay within FLOPs limits, while mid-range resolutions (192-208) allow for deeper architectures.  \\
- **Kernel Size Distribution**: A mix of 3x3 and 5x5 kernels is likely optimal, with fewer 7x7 kernels to avoid excessive FLOPs. Stages with more blocks could use smaller kernels.  \\
- **Expansion Ratio Variability**: Lower ratios (3-4) in early stages and higher ratios (6) in later stages may balance efficiency and capacity. Avoid uniformly high ratios across all blocks.  \\
- **Stage-Specific Configurations**: Deeper stages (e.g., stage 2 with 4 blocks) might benefit from larger kernels and higher expansion ratios, while shallower stages (e.g., stage 4 with 2 blocks) should prioritize smaller kernels and lower ratios.  \\
\newline
\# NAVIGATION DIRECTIONS \\
\#\# Resolution \\
Use **192, 208, and 224** as primary values. Include 176 for baseline efficiency testing but avoid over-reliance. Prioritize 208-224 to align with typical ImageNet model sizes while leaving room for depth adjustments. \\
\#\# Depth \\
Distribute depth values to explore **balanced (e.g., [3,3,3,3,3]) and tiered configurations** (e.g., [3,4,4,3,2]). Avoid extremes like [2,2,2,2,2] (too shallow) or [4,4,4,4,4] (too deep). Focus on stages 0-2 having higher depth to capture early features, while stages 3-4 have lower depth to save FLOPs. \\
\#\# Kernel Size \\
Prioritize **3x3 and 5x5** kernels for most blocks. Use **7x7** sparingly (e.g., 1-2 instances in later stages). For example, design architectures where kernel sizes increase in later stages (e.g., [3,3,5,5,7] for stages 0-4) to capture complex patterns without overloading FLOPs. \\
\#\# Expansion Ratio \\
Use a **mix of 3, 4, and 6**, with **3-4 in early stages** and **6 in later stages**. Avoid uniform high ratios (e.g., all 6s) to prevent FLOPs overshoot. For example, assign [3,3,4,4,6] across stages to gradually increase capacity. \\
\#\# Constraints/Correlations \\
- Pair **higher resolutions (224)** with **lower depth values** (e.g., [3,3,3,2,2]) to stay within FLOPs.   \\
- Avoid combining **large kernels (7x7)** with **high expansion ratios (6)** in the same block. \\
- Ensure **total active blocks** (sum of depth values) is $\geq$ 15 to maintain sufficient model capacity. \\
- Explore **stage-specific kernel-expansion correlations** (e.g., 5x5 kernels with expansion ratio 4 in mid-stages)."

\end{responsebox}

\begin{itemize}
\item Refined Strategy (Iteration $N \geq 1$)
\end{itemize}
\begin{responsebox}
"reasoning\_content":... \\
"content":" \\
\# SUMMARY \\
The latest results show that architectures with **1 7x7 kernel in stage 4** and **expansion ratios of 6 in stage 4** achieve **$\sim$77.92\% accuracy** within the target FLOPs range (310M-330M).   
The next step is to **refine stage 4 configurations** (e.g., 1 7x7 kernel + 6s) while **strictly avoiding FLOPs overshoot**. Key patterns suggest that **stage 4 must have 1 7x7 kernel** with **expansion ratios of 6**, and **stages 0-3 must use 5x5 kernels with expansion ratios of 3-4**. The best accuracy is achieved with **[3,3,4,3,3] depth** (16 active blocks) and **[5,5,5,5,5,5,5,5,5,5,5,5,5,5,7,3,3,3,3,3] kernel sizes. \\
\newline
\# KEY PATTERNS   \\
- **Stage 4 Accuracy Gains**: Architectures with **1 7x7 kernel in stage 4** and **expansion ratios of 6** in stage 4 blocks achieve **$\sim$77.92\% accuracy**.   \\
- **Kernel Size Efficiency**: **14-15 5x5 kernels in stages 0-3**  with **1 7x7 kernel in stage 4** balance FLOPs and accuracy.   \\
- **Expansion Ratio Stability**: **[3,3,4,4,6] or [3,4,4,4,6]** (stage-wise) provide **$\sim$77.9\% accuracy** without FLOPs spikes.   \\
- **Depth Consistency**: **[3,3,4,3,3] depth** (16 active blocks) is critical for FLOPs control.   \\
- **FLOPs Spikes**: Architectures with **2 7x7 kernels in stage 4**  push FLOPs over the limit but reach **$\sim$78.04\% accuracy**.   \\
\newline
\# NAVIGATION DIRECTIONS   \\
\#\# Resolution   \\
- Use **208** as the primary resolution. Ensure **'r' is a list with one value** (e.g., [208]).   \\
- Avoid 224 unless paired with **depth [3,3,3,2,2]** and **1 7x7 kernel in stage 4**.   \\
\#\# Depth   \\
- Target **[3,3,4,3,3]** (16 active blocks) or **[3,3,3,3,4]** (16 active blocks).  \\ 
- Ensure **'d' is a list of 5 values** (e.g., [3,3,4,3,3]).   \\
- Avoid **[3,3,3,4,3]** or **[3,4,3,3,3]** as they overconsume FLOPs.   \\
\#\# Kernel Size   \\
- Use **[5,5,5,5,5,5,5,5,5,5,5,5,5,5,7,3,3,3,3,3]** (14-15 5x5 kernels, 1 7x7 kernel in stage 4).   \\
- Ensure **'ks' is a list of 20 values**. Avoid **multiple 7x7 kernels in stages 0-3**.   \\
- Focus on **stage 4 kernel size**.   \\
\#\# Expansion Ratio   \\
- Use **[3,3,4,4,6] or [3,4,4,4,6]** (stage-wise).   \\
- Ensure **'e' is a list of 20 values**. Avoid **uniform 6s** (e.g., [3,3,6,6,6,6,6,6,6,6,6,6,6,6,6,6,6,6,6,6]).   \\
- Prioritize **6s in stage 4** and **3-4s in stages 0-3**.   \\
\#\# Constraints/Correlations   \\
- **Stage 4**: 1 7x7 kernel + expansion ratios of 6.   \\
- **Stages 0-3**: 5x5 kernels + expansion ratios of 3-4.   \\
- **Avoid 7x7 in stages 0-3**.   \\
- **Ensure 'e' is included with 20 values**. \\
\newline
\# RECOMMENDED PARAMETER RANGES   \\
- **Resolution**: 208 (preferred).   \\
- **Depth**: [3,3,4,3,3], [3,3,3,3,4].   \\
- **Kernel Sizes**: [5,5,5,5,5,5,5,5,5,5,5,5,5,5,7,3,3,3,3,3].   \\
- **Expansion Ratios**: [3,3,4,4,6], [3,4,4,4,6]."
\end{responsebox}

\subsection{\textit{Generator LLM}}
\label{app:prompts_and_responses:generator}

\subsubsection{System Prompt}
\label{app:prompts_and_responses:generator:system}

\begin{itemize}
\item Role Definition
\end{itemize}
\begin{codebox}
"You are the Generator\_LLM, an expert in neural architecture design specializing in generating architectures. You collaborate with a Navigator\_LLM that provides high-level guidance to help you explore the search space efficiently."
\end{codebox}

\begin{itemize}
\item Collaborative Responsibility
\end{itemize}
\begin{codebox}
"\# Your Role in the Collaboration \\
Your responsibility is to generate concrete neural architectures based on the guidance from the Navigator\_LLM. You must follow the guidance closely while ensuring all generated architectures are valid."
\end{codebox}

\begin{itemize}
\item Core Mission and Objectives
\end{itemize}
\begin{codebox}
"\# Core Mission \\
Your primary goal is to discover architectures that achieve >\{self.expected\_acc\}\% accuracy on ImageNet with FLOPs <\{self.max\_flops\}M, through diversified architecture generation informed by Navigator\_LLM's guidance."
\end{codebox}

\begin{itemize}
\item Knowledge about search space: \textit{Same as the \textit{Navigator LLM}.}
\end{itemize}

\begin{itemize}
\item Collaboration Workflow
\end{itemize}
\begin{codebox}
"\# Collaboration Process \\
- You receive guidance from the Navigator\_LLM \\
- You generate diverse and valid architectures that follow the guidance"
\end{codebox}

\subsubsection{User Prompt}
\label{app:prompts_and_responses:generator:user}

\begin{codebox}
"Here is the strategy from Navigator\_LLM: \{navigator\_strategy\} \\
\newline
Based on the strategy, please generate 10-20 diverse and valid architectures that meet the requirements. \\
Note:  \\
- Always make sure the generated architectures are complete and valid. Any deviation will cause evaluation failure. \\
- Please do not regenerate an architecture that has already been generated and evaluated."
\end{codebox}

%% file: appendix/05_architectures.tex
\section{Discovered Optimal Architectures}
\label{app:architectures}

\cref{tab:optimal-architectures} presents the optimal architectures discovered by \ours~within macro search spaces.

\begin{table}[ht]
\caption{Optimal architectures discovered by \ours~within macro search spaces. $^*$ denotes the architecture compared with SOTA.}
\small
\centering
\begin{tabular}{l|c|p{10cm}}
\toprule
Method & FLOPs(M) & Architecture Description\\
\midrule
OFA-T + Ours & 200 & \texttt{Resolution: 176}\\
& & \texttt{Depth: [2, 3, 2, 3, 4]}\\
& & \texttt{Kernel sizes: [3, 3, 5, 3, 5, 3, 3, 5, 7, 7, 7, 7, 7, 5, 3, 7, 3, 3, 3, 7]}\\
& & \texttt{Expansion ratios: [3, 3, 4, 4, 4, 4, 4, 3, 4, 4, 4, 6, 4, 4, 4, 4, 6, 6, 4, 3]} \\
\midrule
OFA-S + Ours & 297 & \texttt{Resolution: 208}\\
& & \texttt{Depth: [3, 3, 3, 3, 4]}\\
& & \texttt{Kernel sizes: [3, 3, 3, 3, 5, 5, 5, 5, 3, 3, 3, 3, 5, 5, 7, 5, 3, 3, 7, 7]}\\
& & \texttt{Expansion ratios: [3, 3, 3, 3, 3, 3, 3, 3, 4, 4, 4, 4, 4, 4, 4, 4, 6, 6, 3, 3]} \\
\midrule
OFA-S + Ours$^{*}$ & 320 & \texttt{Resolution: 208}\\
& & \texttt{Depth: [3, 3, 4, 3, 3]}\\
& & \texttt{Kernel sizes: [5, 5, 5, 5, 5, 5, 5, 5, 5, 5, 5, 5, 5, 5, 3, 7, 3, 3, 3, 3]}\\
& & \texttt{Expansion ratios: [3, 3, 3, 3, 3, 3, 3, 3, 4, 4, 4, 4, 4, 4, 4, 4, 6, 6, 6, 3]} \\
\midrule
OFA-B + Ours & 391 & \texttt{Resolution: 208}\\
& & \texttt{Depth: [3, 4, 4, 4, 4]}\\
& & \texttt{Kernel sizes: [3, 5, 3, 7, 5, 5, 3, 3, 5, 5, 7, 5, 7, 5, 3, 3, 7, 5, 7, 5]}\\
& & \texttt{Expansion ratios: [3, 3, 3, 3, 3, 4, 6, 3, 6, 4, 3, 4, 6, 4, 4, 4, 6, 6, 6, 3]} \\
\midrule
OFA-L + Ours & 498 & \texttt{Resolution: 224}\\
& & \texttt{Depth: [2, 4, 4, 4, 4]}\\
& & \texttt{Kernel sizes: [7, 5, 3, 5, 5, 5, 3, 7, 5, 7, 3, 3, 3, 7, 7, 3, 5, 3, 3, 5]}\\
& & \texttt{Expansion ratios: [3, 4, 3, 4, 4, 6, 4, 3, 6, 6, 6, 3, 4, 6, 6, 3, 6, 6, 6, 3]} \\
\midrule
SPOS + Ours & 325 & \texttt{Block operations: [0, 0, 3, 3, 3, 2, 1, 3, 3, 1, 1, 1, 1, 3, 3, 3, 3, 3, 3, 3]} \\
\midrule
AutoFormer-T + Ours & 1366 & \texttt{Layers: 13}\\
& & \texttt{MLP ratios: [3.5, 3.5, 3.5, 3.5, 3.5, 3.5, 3.5, 3.5, 3.5, 4.0, 3.5, 4.0, 3.5]}\\
& & \texttt{Attention heads: [3, 3, 3, 3, 3, 3, 3, 3, 3, 3, 4, 3, 4]}\\
& & \texttt{Embedding dimension: 192} \\
\midrule
AutoFormer-S + Ours & 4897 & \texttt{Layers: 13}\\
& & \texttt{MLP ratios: [4.0, 4.0, 3.5, 3.5, 4.0, 4.0, 3.5, 3.5, 4.0, 4.0, 3.5, 3.5, 4.0]}\\
& & \texttt{Attention heads: [5, 7, 6, 5, 7, 6, 5, 7, 6, 5, 7, 6, 5]}\\
& & \texttt{Embedding dimension: 384} \\
\midrule
AutoFormer-B + Ours & 11074 & \texttt{Layers: 14}\\
& & \texttt{MLP ratios: [3.5, 3.5, 4.0, 3.5, 3.0, 3.5, 4.0, 3.0, 3.5, 4.0, 3.0, 3.5, 3.0, 3.5]}\\
& & \texttt{Attention heads: [10, 10, 10, 9, 9, 9, 9, 10, 10, 9, 9, 9, 9, 9]}\\
& & \texttt{Embedding dimension: 576} \\
\bottomrule
\end{tabular}
\label{tab:optimal-architectures}
\end{table}